\PassOptionsToPackage{dvipsnames,table}{xcolor}
\documentclass[10pt,twocolumn,letterpaper]{article}

\usepackage[pagenumbers]{cvpr}

\definecolor{cvprblue}{rgb}{0.21,0.49,0.74}
\usepackage[pagebackref,breaklinks,colorlinks,allcolors=cvprblue]{hyperref}

\usepackage{amsmath}
\usepackage{multirow}
\usepackage{multicol}
\usepackage{tikz}
\usepackage[ruled,lined,linesnumbered]{algorithm2e}
\usepackage{enumitem}

\newcommand{\goldtext}[1]{{\textcolor{YellowOrange}{\textbf{#1}}}\xspace}
\newcommand{\silvertext}[1]{{\textcolor{Gray}{\textbf{#1}}}\xspace}

\ifx\figurename\undefined \def\figurename{Figure}\fi
\renewcommand{\figurename}{Fig.}
\renewcommand{\paragraph}[1]{\textit{\textbf{#1}} }

\newcommand{\Sect}[1]{Sec.~\ref{#1}}
\newcommand{\Fig}[1]{Fig.~\ref{#1}}
\newcommand{\Tbl}[1]{Tbl.~\ref{#1}}

\newcommand{\specialcell}[2][c]{\begin{tabular}[#1]{@{}c@{}}#2\end{tabular}}

\definecolor{inkgreen}{HTML}{0B5D3B}
\newcommand{\best}[1]{\textbf{\textcolor{inkgreen}{#1}}}

\newcommand{\proj}{\textsc{Nova}\xspace}

\newcommand{\mode}[1]{\underline{\textsc{#1}}\xspace}

\graphicspath{{figs/}}

\def\confName{CVPR}
\def\confYear{2027}

\title{\textsc{No\raisebox{\depth}{\scalebox{1}[-1]{a}}a}\xspace: \underline{N}ormalization-Aware \underline{O}utput-Guided \underline{A}ttention \underline{A}cceleration for Video Diffusion Transformers}
\author{
  Wenxuan Miao$^{1}$,
  Haosong Liu$^{1}$,
  Aiyue Chen$^{3}$,
  Yiwu Yao$^{3}$,\\
  Zihan Liu$^{1}$,
  Anbang Wu$^{1}$,
  Yiming Gan$^{4}$,
  Jingwen Leng$^{1, 2}$,
  Minyi Guo$^{1, 2}$,
  Yu Feng$^{1,2}$\thanks{Corresponding author.}\\
  $^{1}$Shanghai Jiao Tong University, \quad
  $^{2}$Shanghai Qizhi Institute,\quad \\
  $^{3}$Huawei Technologies Co.,Ltd,\quad
  $^{4}$Institute of Computing Technology, Chinese Academy of Science
}

\begin{document}
\maketitle

\begin{abstract}
Video diffusion transformers (vDiTs) have become the main backbone for video generation.
The quadratic cost of self-attention limits their efficiency at high resolutions and long durations.
Existing sparse attention methods often rely on preserving high attention weights.
However, these criteria do not fully capture the fidelity of the final value-weighted output, whose error depends jointly on changes to the attention numerator and the softmax denominator.

We propose \proj, a coarse-to-fine sparse attention framework that first approximates the full normalized output using size-weighted key/value centroids.
Building on this estimate, we construct a residual proxy that accounts for coupled numerator-denominator errors by centering value contributions around the estimated output.
This proxy guides the selection of query-key blocks for refinement under a predefined block budget for each query cluster.
Selected blocks are evaluated through exact token-level computation, while the remaining blocks contribute through their centroids under a shared softmax normalization.
On five vDiTs, \proj achieves the best PSNR, SSIM, and LPIPS among sparse attention methods, improving PSNR by up to 1.55\,dB over the strongest baseline, while delivering up to 1.82$\times$ end-to-end speedup on an H100 GPU.
\end{abstract}

\section{Introduction}
Generating temporally consistent and photorealistic videos is an important step toward building world models that understand the physical world~\cite{brooks2024video, bruce2024genie, yang2024unisim, agarwal2025cosmos}.
This requires the model to possess the ability to fuse spatio-temporal information across all frames.
Video diffusion transformers~\cite{peebles2023dit, yang2025cogvideox, zheng2024opensora, kong2024hunyuanvideo, wan2025wan, longcat2025video, wan2025wan22} use global self-attention to aggregate information across video tokens, but dense attention incurs a quadratic cost as the sequence grows longer.
This scaling makes self-attention a main optimization target for efficient video generation~\cite{xi2025sparse, yang2026sparse, zhang2025sta, zhang2025spargeattn, luo2026svoo, zhou2026svg}.

\begin{figure}[t]
    \centering
    \includegraphics[width=\linewidth]{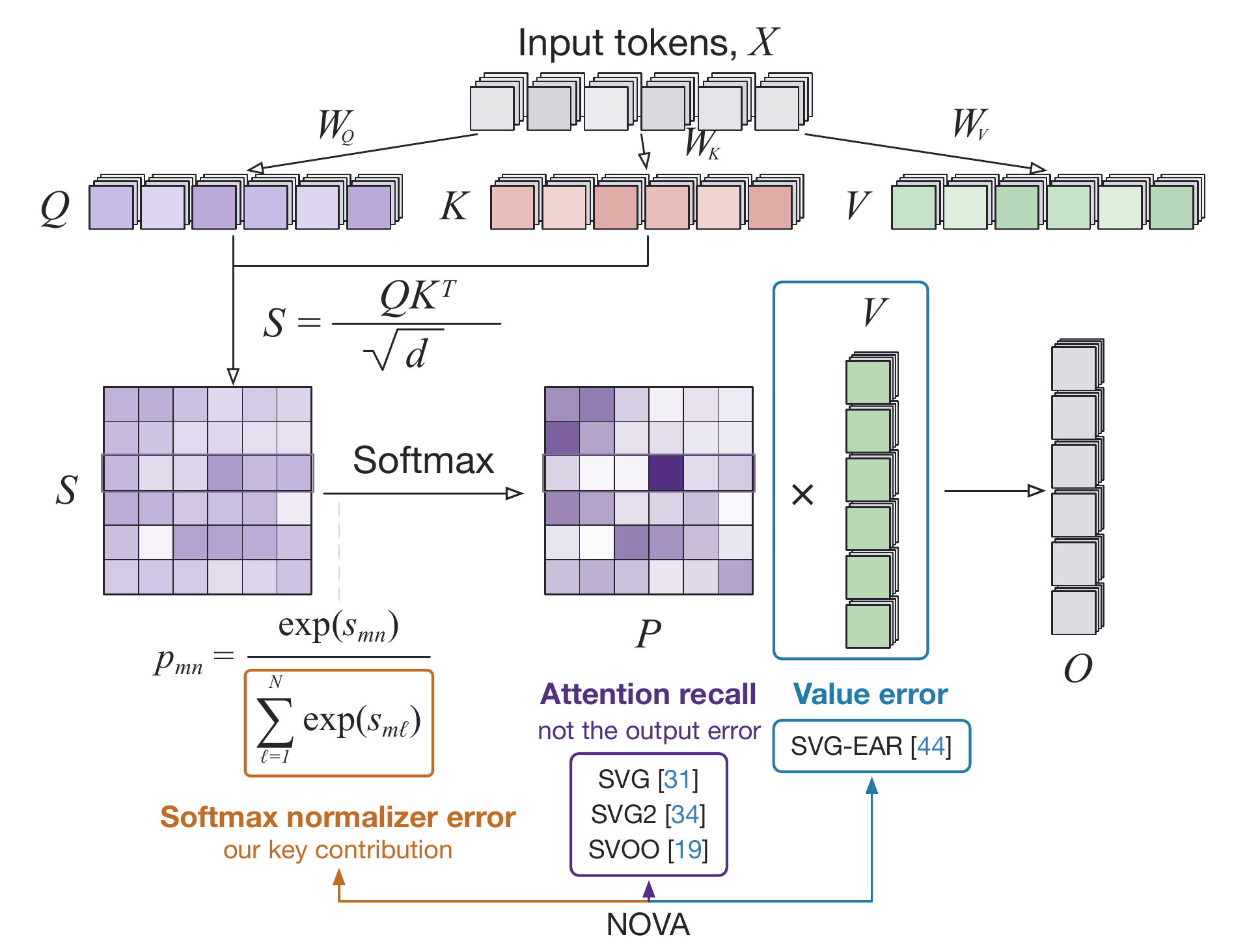}
    \caption{
        Attention computation and two components relevant to sparse-output error.
        The blue box highlights value aggregation, while the orange box
        highlights the softmax normalizer.
    }
    \label{fig:attention_mechanism}
\end{figure}

Recent methods~\cite{xi2025sparse, yang2026sparse, luo2026svoo, zhou2026svg} primarily adopt sparse attention to accelerate video generation.
It reduces computational cost by computing attention only for selected query–key pairs.
The key question is how to measure importance and decide which computations to retain.
A common assumption in sparse attention is that larger attention weights indicate greater importance.
Following this assumption, existing methods seek high attention recall as a proxy for output fidelity.
They build a sparse mask that keeps only query–key pairs with large estimated attention weights and skips the rest.
To do so, SVG~\cite{xi2025sparse} assigns each attention head a static pattern, whereas SVG2~\cite{yang2026sparse} and SVOO~\cite{luo2026svoo} dynamically cluster queries and keys into blocks and retain the blocks with high attention weights.

\paragraph{Issues.}
However, high attention recall alone does not guarantee a low attention-output mean-squared error (MSE): recall only counts the retained attention mass, while the output error is determined by both the \textit{discarded values} and the \textit{softmax renormalization}.
As shown in \Fig{fig:attention_mechanism}, each attention weight, $P$, scales its associated values, $V$, during value aggregation (highlighted in blue box), so equal weights can make different contributions to the output.
Discarding attention weights both removes their value contributions and changes the softmax normalizer (highlighted in orange box), thereby rescaling all retained contributions.
While prior work~\cite{xi2025sparse, yang2026sparse, luo2026svoo, zhang2025spargeattn} primarily targets high attention recall, recall alone cannot capture these mixture effects.
This key observation motivates us to propose a sparse selection strategy based on normalized attention-output error, rather than attention recall alone.

Some recent methods go beyond attention weights and account for value vectors, but their scoring objectives still omit the effects of softmax renormalization, potentially misestimating the resulting output error.
For example, SVG-EAR~\cite{zhou2026svg} estimates the error of centroid-based value compensation to select blocks for exact computation.
However, its selection score is computed from unnormalized value contributions, so it does not account for how discarding blocks rescales the retained ones through the shared softmax denominator.

\paragraph{Idea.}
To account for both effects, we select blocks by their contribution to the normalized attention-output error rather than by their attention weights alone.
Our analysis shows that, after softmax renormalization, the error from discarding a block is determined by its attention weights multiplied by the differences between its value vectors and the attention output.
Consequently, a block with large attention weights contributes little error if its values are close to the output, whereas a block with moderate weights but distant values can dominate the error.

We propose \proj, a coarse-to-fine sparse attention method.
\proj works in four steps. 
First, it computes the coarse output from all key/value cluster centroids, weighting each centroid by the number of tokens in its cluster. 
Second, it scores each block by how much dropping it would change the coarse output, relative to its computation cost. 
Third, it selects blocks for each query cluster: top-$p$ coverage of the centroid attention weights sets a token-pair budget, and the highest-scoring blocks fill this budget.
Finally, it computes the selected blocks exactly and approximates the remaining blocks by their centroids, with both parts sharing the same softmax normalization to produce the final output. 
As a result, \proj preserves higher generation quality than prior work, by up to 1.55~dB PSNR (\Sect{sec:eval}).

\paragraph{Results.}
We evaluate \proj on five widely used video diffusion transformers~\cite{kong2024hunyuanvideo, wan2025wan, wan2025wan22, lingbot2026video, longcat2025video}, all generating 720p videos.
We compare against the state-of-the-art sparse attention methods, SVG~\cite{xi2025sparse}, 	SVG2~\cite{yang2026sparse}, SVOO~\cite{luo2026svoo}, and SVG-EAR~\cite{zhou2026svg}, measuring video quality with VBench~\cite{huang2024vbench} and fidelity to dense attention with PSNR, SSIM~\cite{wang2004ssim}, and LPIPS~\cite{zhang2018lpips}, and timing all methods on a single NVIDIA H100 GPU.
\proj achieves the best PSNR, SSIM, and LPIPS among sparse methods on all five models, improving PSNR over the strongest baseline by up to 1.55~dB, and obtains the highest VBench score among sparse methods on four of the five models.
Meanwhile, \proj delivers up to 1.82$\times$ end-to-end speedup over dense attention, the fastest on four of the five models and within 0.01$\times$ of the fastest on LongCat-Video.
Our contributions are listed as follows:
\begin{itemize}
    \item We derive the exact dependence of attention-output MSE on coupled numerator--denominator perturbations and use the resulting centered value residual to construct a normalization-aware block-scoring proxy.
    \item We develop a complete-output estimation and selective-refinement procedure that uses this proxy to allocate exact block computation, combining exact interactions and centroid estimates under one normalization.
    \item On five video diffusion transformers, \proj achieves the best fidelity among sparse methods, improving PSNR by up to 1.55~dB, with up to 1.82$\times$ end-to-end speedup.
\end{itemize}
\section{Related Work}

\paragraph{Efficient Video Diffusion.}
Video diffusion models apply a large network over many denoising steps~\cite{ho2022video,kong2024hunyuanvideo,wan2025wan}, so acceleration reduces either the number of steps or the cost per step.
Fast solvers~\cite{song2021ddim,lu2022dpmsolver} and distillation~\cite{salimans2022progressive} reduce steps, while attention broadcasting~\cite{zhao2025pab} and feature caching~\cite{zou2025toca,liu2026astraea} reuse computation across steps.
Within each step, token merging~\cite{bolya2023tome,bolya2023tomesd} compresses redundant features, and sparse attention~\cite{xi2025sparse,yang2026sparse,luo2026svoo,zhou2026svg,zhang2025sta,zhang2025spargeattn} skips unimportant token interactions.

\paragraph{Clustered Sparse Attention.}
In language models, sparse attention uses static patterns~\cite{child2019sparse,beltagy2020longformer,zaheer2020bigbird,xiao2024streamingllm}, content-based grouping~\cite{kitaev2020reformer,roy2021routing}, or dynamic block selection~\cite{tang2024quest,jiang2024minference,yuan2025nsa,lu2025moba}.
Video diffusion transformers share the quadratic cost but have spatial--temporal attention structure, motivating video-specific designs.
SVG~\cite{xi2025sparse} assigns each head a predefined spatial or temporal pattern.
SVG2~\cite{yang2026sparse} instead clusters queries and keys by similarity into contiguous blocks and selects blocks by estimated attention mass.
Rather than clustering them independently, SVOO~\cite{luo2026svoo} co-clusters queries and keys by their attention profiles.
We adopt this co-clustering, but select blocks by their estimated effect on the normalized output rather than by attention mass.

\paragraph{Error-aware Centroid Compensation.}
\mode{SVG-EAR}~\cite{zhou2026svg} compensates unselected blocks with centroids and ranks blocks by unnormalized approximation error per unit cost.
\mode{\proj} instead accounts for the shared softmax denominator when scoring their effects on the output.
\section{Methodology}
\label{sec:method}
Our goal is to reduce attention computation while preserving the normalized attention output.
We organize \proj as a coarse-to-fine sparse attention method: a cluster-level output estimate guides block selection, and exact computation on selected blocks refines the approximation.
The final output combines exact and centroid-based contributions under a shared softmax normalization.

\subsection{Preliminary}
\label{sec:method_preliminary}
We summarize standard normalized self-attention and define the cluster and block-budget notation used in the subsequent analysis and algorithm.

The softmax denominator jointly aggregates the contributions of all keys, so we express attention as a ratio of two sums. We consider a single attention head with $L$ tokens and queries, keys, and values ($Q,K,V\in\mathbb R^{L\times d}$), omitting batch, layer, head, and diffusion-step indices. For query $q_m$,
\begin{equation}
    \begin{aligned}
        a_{mn} &= \exp\!\left(q_m^\top k_n/\sqrt{d}\right),
        \qquad N_m = \sum_{n=1}^{L} a_{mn}v_n,\\
        Z_m &= \sum_{n=1}^{L} a_{mn},
        \qquad O_m = N_m/Z_m.
    \end{aligned}
    \label{eq:method_attention}
\end{equation}
Here, $a_{mn}$ is an unnormalized attention weight, $N_m$ is the weighted value sum, and $Z_m$ is softmax denominator.
We use $m, n$ for token indices and $i, j$ for cluster indices; bars are cluster centroids and hats are centroid-based estimates.

\begin{figure*}[t]
    \centering
    \includegraphics[width=\textwidth]{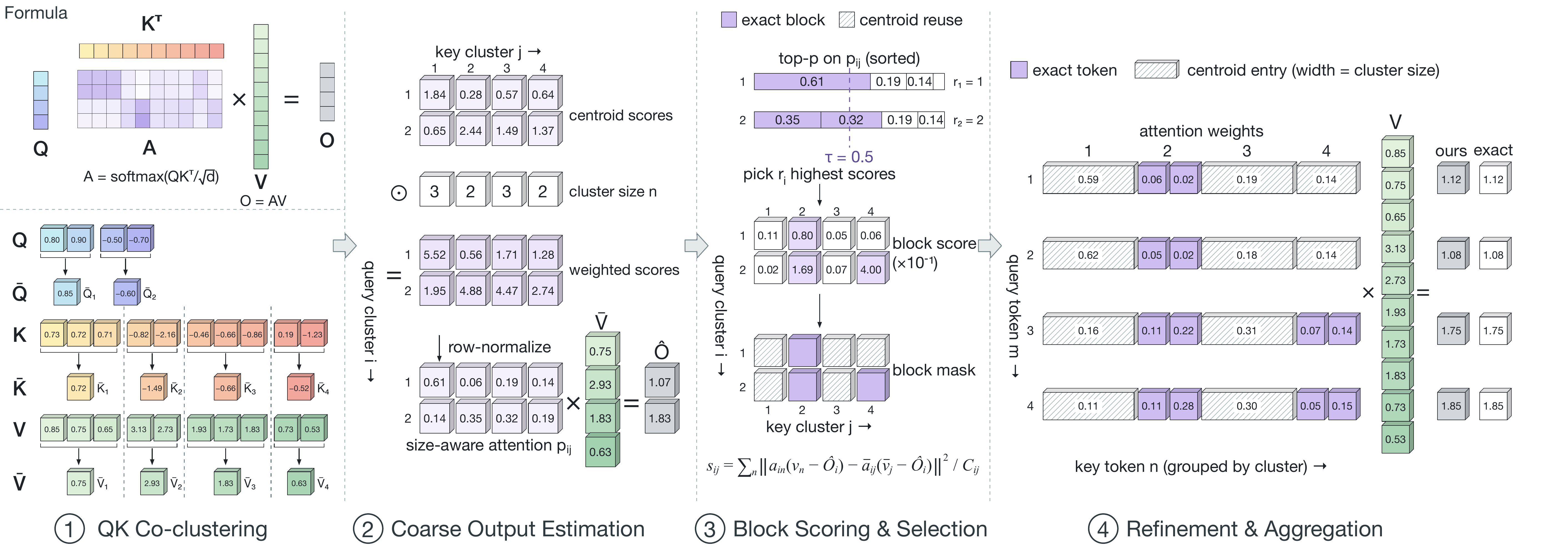}
    \caption{\textbf{Overview of \proj}($d{=}2$, 4 queries, 10 keys).
    \textcircled{\scriptsize 1}~QK co-clustering partitions queries and keys into clusters $\mathcal{Q}_i$ and $\mathcal{K}_j$, summarized by centroids $\bar q_i$, $\bar k_j$, $\bar v_j$ (Eq.~\eqref{eq:centroids}).
    \textcircled{\scriptsize 2}~Size-weighted centroid attention $p_{ij}$ gives the coarse output $\hat O_i$ of each query cluster (Eq.~\eqref{eq:method_coarse_output}).
    \textcircled{\scriptsize 3}~Each block is scored by its estimated contribution to the normalized output error $s_{ij}$ (Eq.~\eqref{eq:method_block_score}).
    In this illustrative example, top-$p$ determines the number of blocks selected for each query cluster; the full method uses a token-pair budget.
    \textcircled{\scriptsize 4}~Selected blocks are computed exactly at token level, while each unselected block is approximated by a single centroid term. Both parts are summed in the same softmax numerator and denominator (Eq.~\eqref{eq:method_refined_output}).}
    \label{fig:overview}
\end{figure*}

\subsection{Overview}
\label{sec:overview}
\Fig{fig:overview} illustrates the workflow of \proj, which processes each attention head in four stages.
First, \proj co-clusters queries and keys with similar attention behavior, reorders them into contiguous blocks, and computes a coarse output for each query cluster from size-weighted key/value centroids (\Sect{sec:coarse}).
Second, it uses this coarse output to score each query-key block by its output error, accounting for both value approximation and softmax renormalization (\Sect{sec:analysis}).
Third, it sets a token-pair budget for each query cluster from the top-$p$ coverage of the centroid attention weights, and fills this budget with the top-scoring blocks (\Sect{sec:selection}).
Finally, it computes the selected blocks exactly and approximates the remaining ones by their centroids, combining both parts under a shared softmax normalization (\Sect{sec:refine}).

\subsection{Coarse Output Estimation}
\label{sec:coarse}
\begin{figure}[t]
    \centering
    \includegraphics[width=\columnwidth]{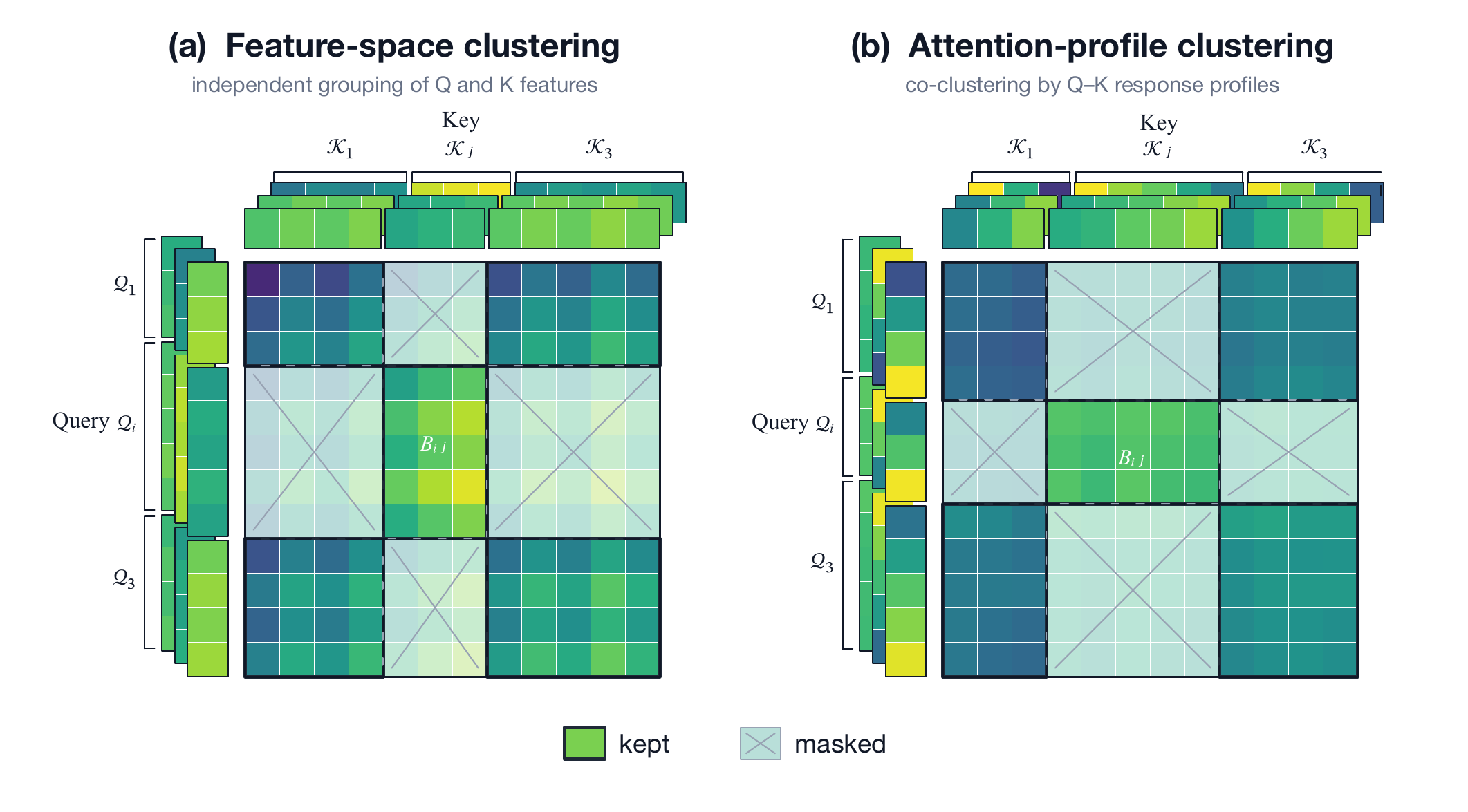}
    \caption{\textbf{Feature-space \vs attention-profile clustering.}
    (a)~Clustering queries and keys separately by feature similarity~\cite{yang2026sparse} produces blocks $B_{ij}$ with mixed attention scores, so skipping a block also removes important entries.
    (b)~Co-clustering~\cite{luo2026svoo} by attention profiles (Eq.~\eqref{eq:method_clustering}) groups tokens with similar attention patterns, so each block is either uniformly important or uniformly negligible.}
    \label{fig:clustering}
\end{figure}

\paragraph{QK Co-clustering.}
Here, we follow the block-based sparse attention paradigm, which is widely adopted in prior work~\cite{jiang2024minference, zhang2025spargeattn, yang2026sparse, luo2026svoo, zhou2026svg}.
Block-based sparse attention is compatible with existing GPU SIMT execution and more efficient than unstructured token-pair sparsity~~\cite{gray2017blocksparse, dao2022flashattention}.

To form such blocks, we cluster and reorder queries and keys into contiguous blocks of tokens with similar attention patterns, which also enables block-level selection and centroid-based output estimation.
We use QK co-clustering~\cite{luo2026svoo}, which partitions the $L$ queries into $Q_C$ clusters $\{\mathcal{Q}_i\}_{i=1}^{Q_C}$ and the $L$ keys into $K_C$ clusters $\{\mathcal{K}_j\}_{j=1}^{K_C}$, with sizes $n^Q_i = |\mathcal{Q}_i|$ and $n^K_j = |\mathcal{K}_j|$.
Each cluster is summarized by the mean of its members,
\begin{equation}
    \bar q_i = \frac{1}{n^Q_i}\sum_{m\in\mathcal{Q}_i} q_m, \quad
    \bar k_j = \frac{1}{n^K_j}\sum_{n\in\mathcal{K}_j} k_n, \quad
    \bar v_j = \frac{1}{n^K_j}\sum_{n\in\mathcal{K}_j} v_n,
    \label{eq:centroids}
\end{equation}
and the centroids are stacked into $\bar Q = [\bar q_1, \dots, \bar q_{Q_C}]^\top \in \mathbb{R}^{Q_C \times d}$ and $\bar K = [\bar k_1, \dots, \bar k_{K_C}]^\top \in \mathbb{R}^{K_C \times d}$.
Values share the key partition, so $\bar v_j$ is the value centroid of key cluster $\mathcal{K}_j$.
Instead of clustering tokens by their raw features, we represent each token by its attention scores against the centroids of the other side: a query $q_m$ is represented by $\bar K q_m \in \mathbb{R}^{K_C}$, its dot products with all key centroids, and a key $k_n$ by $\bar Q k_n \in \mathbb{R}^{Q_C}$.
Each token is then compared with every candidate cluster on its own side using the cosine similarity between these attention-score vectors:
\begin{equation}
    \begin{aligned}
        \operatorname{sim}_Q(m,i) &= \cos\!\left(\bar K q_m,\ \bar K \bar q_i\right),\\
        \operatorname{sim}_K(n,j) &= \cos\!\left(\bar Q k_n,\ \bar Q \bar k_j\right).
    \end{aligned}
    \label{eq:method_clustering}
\end{equation}
Here, $\operatorname{sim}_Q(m,i)$ measures how similarly query $m$ and query centroid $\bar q_i$ attend to the key centroids, and $\operatorname{sim}_K(n,j)$ measures how similarly key $n$ and key centroid $\bar k_j$ are attended to by the query centroids.
The scaling factor $1/\sqrt{d}$ is omitted because cosine similarity is scale-invariant.
Each query is assigned to the cluster with the highest $\operatorname{sim}_Q(m,i)$, and each key to the cluster with the highest $\operatorname{sim}_K(n,j)$.
We alternate between updating the query clusters with $\bar K$ fixed and updating the key clusters with $\bar Q$ fixed, recomputing the centroids after each update.

Repeated clustering introduces overhead across diffusion steps.
To reduce this cost, we retain the Q/K centroids separately for each layer and attention head.
Since the attention behavior of tokens changes only slightly between adjacent denoising steps, we warm-start clustering from the centroids of the previous sparse step, which keeps the clusters accurate with only two iterations per step.
Queries and key/value pairs are then reordered by their cluster assignments to form contiguous blocks $B_{ij}=\mathcal{Q}_i\times\mathcal{K}_j$.

\paragraph{Coarse Attention Output.}
Following centroid compensation~\cite{zhou2026svg}, we approximate the full attention output without evaluating all token pairs by computing attention between query centroids and all key/value centroids.
For query cluster $i \in \{1,\dots,Q_C\}$ and key cluster $j \in \{1,\dots,K_C\}$, let $\bar a_{ij} = \exp(\bar q_i^\top \bar k_j/\sqrt{d})$ denote the unnormalized centroid attention weight.
Each key/value centroid represents an entire cluster of $n_j^K$ tokens, so we approximate the total weight of the cluster by $\sum_{n\in\mathcal{K}_j} a_{mn} \approx n_j^K \bar a_{ij}$ for every query $m \in \mathcal{Q}_i$.
This gives the coarse weighted value sum $\hat N_i$, the coarse softmax denominator $\hat Z_i$, and the coarse output $\hat O_i$ of query cluster $i$:
\begin{equation}
    \begin{aligned}
        \hat N_i &= \sum_{j=1}^{K_C} n_j^K \bar a_{ij}\, \bar v_j,
        \qquad
        \hat Z_i = \sum_{j=1}^{K_C} n_j^K \bar a_{ij},\\
        p_{ij} &= \frac{n_j^K \bar a_{ij}}{\hat Z_i},
        \qquad
        \hat O_i = \frac{\hat N_i}{\hat Z_i} = \sum_{j=1}^{K_C} p_{ij}\, \bar v_j.
    \end{aligned}
    \label{eq:method_coarse_output}
\end{equation}

Here, $p_{ij}$ is the estimated attention mass that query cluster $i$ assigns to key cluster $j$, with $\sum_j p_{ij} = 1$, and $\hat O_i$ serves as a shared estimate of the attention output for all queries in $\mathcal{Q}_i$.
The factor $n_j^K$ accounts for the number of tokens represented by each centroid in both the weighted value sum and the softmax denominator.
We next analyze how refining individual blocks changes this coarse output.

\begin{figure*}[t]
    \centering
    \includegraphics[width=\textwidth]{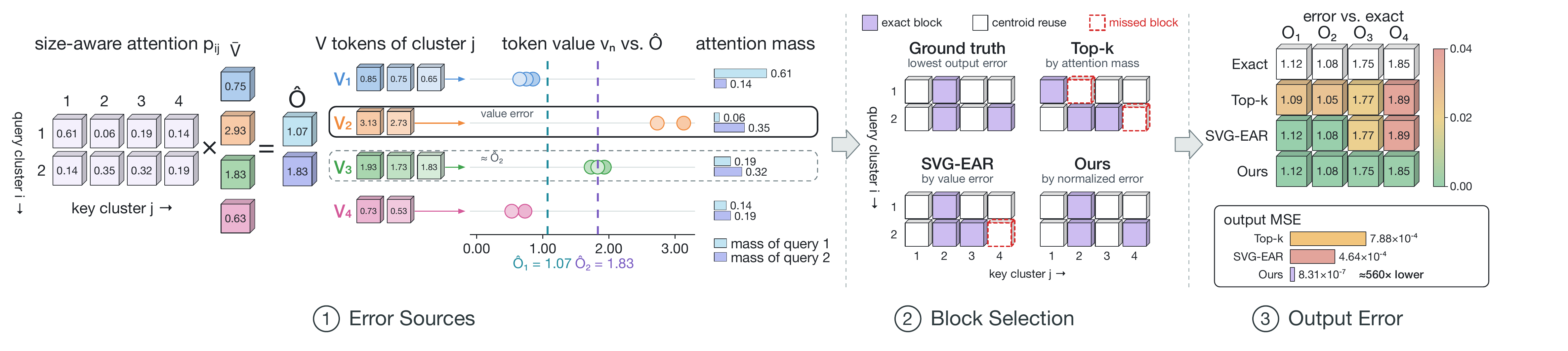}
    \caption{\textbf{Block selection by attention weight \vs output error} on the example of \Fig{fig:overview}. Blocks with high attention weight are not necessarily the ones whose centroid approximation causes large error. Top-$k$ selects by weight and SVG-EAR~\cite{zhou2026svg} ignores softmax renormalization, so both keep the wrong blocks. \proj selects by normalized output error and lowers the output MSE by about $560\times$.}
    \label{fig:comparison}
\end{figure*}
\subsection{Normalized Output Error Analysis}
\label{sec:analysis}
\paragraph{Effect of Block Refinement.}
Replacing a centroid approximation with exact computation changes both the weighted value sum and the softmax denominator.
A block score should therefore account for their combined effect on the normalized output.
At query centroid $\bar q_i$, let $\hat N_i$ and $\hat Z_i$ be the coarse weighted value sum and softmax denominator from Eq.~\eqref{eq:method_coarse_output}.
Consider replacing the centroid approximation of block $j$ with exact computation while keeping all other blocks unchanged.
Let $\Delta N_{ij}$ and $\Delta Z_{ij}$ denote the differences between the exact and centroid contributions of this block
to the numerator and denominator, respectively.
The resulting output change is
\begin{equation}
    \begin{aligned}
        \Delta O_{ij}
        &= \frac{\hat N_i+\Delta N_{ij}}{\hat Z_i+\Delta Z_{ij}}
        -\frac{\hat N_i}{\hat Z_i}
        = \frac{\Delta N_{ij}-\hat O_i\Delta Z_{ij}}
               {\hat Z_i+\Delta Z_{ij}}\\
        &\approx \frac{\Delta N_{ij}-\hat O_i\Delta Z_{ij}}{\hat Z_i}.
    \end{aligned}
    \label{eq:method_output_change}
\end{equation}
The last expression is a first-order approximation for small
$|\Delta Z_{ij}|/\hat Z_i$.
The term $-\hat O_i\Delta Z_{ij}$ accounts for the output change caused by softmax renormalization.
We empirically evaluate the accuracy of this first-order approximation across different video diffusion models in the supplementary material.

\paragraph{Value Contributions and Normalization.}
Eq.~\eqref{eq:method_output_change} shows that the effect of refining a block depends not only on its numerator and denominator changes, but also on the current output estimate $\hat O_i$.
This dependence arises because changing the softmax denominator rescales the contributions of all blocks.
We will make this dependence explicit by expressing $\Delta N_{ij}-\hat O_i\Delta Z_{ij}$ as a sum of token-level residuals.

For $n\in\mathcal{K}_j$, let $a_{in}=\exp(\bar q_i^\top k_n/\sqrt{d})$ be the token-level attention weight evaluated at query centroid $\bar q_i$.
Replacing the centroid approximation changes the token's numerator contribution by $a_{in}v_n-\bar a_{ij}\bar v_j$ and its denominator contribution by $a_{in}-\bar a_{ij}$.
Substituting these changes gives
\begin{equation}
    \begin{aligned}
        r_{ijn}
        &=
        a_{in}(v_n-\hat O_i)
        -\bar a_{ij}(\bar v_j-\hat O_i),\\
        \Delta N_{ij}-\hat O_i\Delta Z_{ij}
        &=
        \sum_{n\in\mathcal{K}_j}r_{ijn}.
    \end{aligned}
    \label{eq:method_centered_residual}
\end{equation}
Centering both the original and centroid values at $\hat O_i$ therefore accounts for value approximation and softmax renormalization together.
As illustrated in \Fig{fig:comparison}\,\textcircled{\scriptsize 1}, a block with small attention mass but values far from $\hat O_i$ can dominate the error, whereas a block with large mass but values close to $\hat O_i$ contributes little.

\subsection{Block Scoring and Selection}
\label{sec:selection}
\paragraph{Block Scoring.}
Block selection requires a score that can be evaluated independently for each candidate.
We use the sum of squared token residuals as an additive error proxy and divide it by the block's token-pair cost $C_{ij}=n_i^Qn_j^K$:
\begin{equation}
    s_{ij}
    =
    \frac{1}{C_{ij}}
    \sum_{n\in\mathcal{K}_j}
    \lVert r_{ijn}\rVert_2^2.
    \label{eq:method_block_score}
\end{equation}
The factor $\hat Z_i^{-2}$ is shared within query cluster $i$ and does not affect block rankings.
The score prioritizes blocks with larger estimated errors relative to their computation cost.
The supplementary material derives the joint first-order output error of the unselected blocks and provides an upper-bound interpretation of the block score in Eq.~\eqref{eq:method_block_score}.

\paragraph{Block Selection.}
The number of exact blocks should adapt to the concentration of the estimated attention distribution.
For each query cluster, we sort $p_{ij}$ in descending order and accumulate the weights until their sum first exceeds a top-$p$ threshold~\cite{holtzman2020curious, yang2026sparse} $\tau\in(0,1)$.
The token-pair area of the accumulated key blocks defines the budget $R_i=\sum_{j\in\mathcal{J}_i^{\text{top-}p}} C_{ij}$.
We then select blocks in descending order of the score in Eq.~\eqref{eq:method_block_score} until the budget $R_i$ is filled, denoting their key-cluster indices by $\mathcal{J}_i$.
Thus, the estimated attention weights determine how large a budget is spent, while the error scores determine which blocks are selected.

\begin{figure}[t]
    \centering
    \includegraphics[width=\columnwidth]{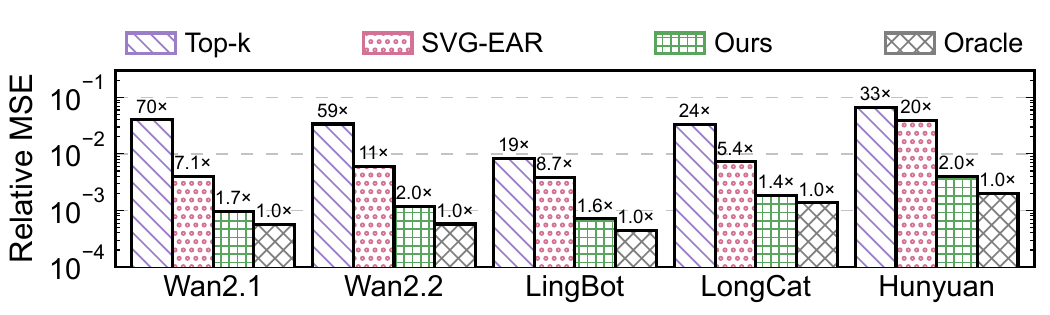}
    \caption{\textbf{Attention output error on real models.} Error of Top-$k$, SVG-EAR, and \proj against full attention on five video models, with all methods keeping the same share of attention. \mode{Oracle} greedily selects blocks using residual scores computed from exact token-level contributions and dense-attention outputs under the same target token-pair budget as \mode{\proj}.}
    \label{fig:mse}
\end{figure}

\paragraph{Error on Real Models.}
\Fig{fig:mse} compares the attention output error of block \mode{Top-$k$}, \mode{SVG-EAR}, and \mode{\proj} against exact attention on five video models at equal attention density.
We report the geometric mean of the relative MSE over 120 evenly spaced layers and denoising steps.
\mode{Oracle} computes block residuals from exact token-level contributions and dense-attention outputs.
\mode{\proj} has the lowest error among practical methods: $4$--$10\times$ lower than \mode{SVG-EAR}, $12$--$41\times$ lower than \mode{Top-$k$}, and within $2\times$ of \mode{Oracle}.
Scoring value error without renormalization thus only partly closes the gap, consistent with \Sect{sec:analysis}.
\subsection{Output Refinement}
\label{sec:refine}
Unselected blocks still contribute to both the output and its softmax denominator.
We retain their centroid approximations and combine them with exact contributions from the selected blocks under a shared normalization.
For query $q_m\in\mathcal{Q}_i$, let $b_{mj}=n_j^K\exp(q_m^\top\bar k_j/\sqrt{d})$ be the size-weighted centroid attention weight.
The final output is
\begin{equation}
    \tilde O_m
    =
    \frac{
        \displaystyle
        \sum_{j\in\mathcal{J}_i}
        \sum_{n\in\mathcal{K}_j}a_{mn}v_n
        +
        \sum_{j\notin\mathcal{J}_i}b_{mj}\bar v_j
    }{
        \displaystyle
        \sum_{j\in\mathcal{J}_i}
        \sum_{n\in\mathcal{K}_j}a_{mn}
        +
        \sum_{j\notin\mathcal{J}_i}b_{mj}
    }.
    \label{eq:method_refined_output}
\end{equation}
Each selected block replaces its centroid contribution in both the numerator and denominator.
The resulting outputs are restored to the original query order.

\subsection{Efficiency Analysis}

\paragraph{Kernel Implementation.}
Storing residual vectors for each query-centroid/key-token pair would introduce substantial memory traffic.
We instead use a Triton~\cite{tillet2019triton} scoring kernel that fuses token-level score evaluation and reduction, accumulating block scores without materializing the residual vectors.
Value statistics are precomputed and reused across query clusters.
Selected blocks are evaluated using FlashInfer's~\cite{ye2025flashinfer} variable-block sparse attention backend.
A Triton fusion kernel combines the exact sparse output and its log-sum-exp~\cite{dao2022flashattention} statistics with the remaining centroid contributions to implement the shared normalization in Eq.~\eqref{eq:method_refined_output}.

\paragraph{Computational Cost.}
The additional work must remain small relative to token-level attention for sparsification to be useful.
After clustering and token reordering, coarse output estimation operates on centroid pairs, block scoring compares query centroids with key tokens, and the final centroid approximation pairs queries with key centroids.
Including row-wise sorting, their combined arithmetic cost is
\begin{equation}
    \begin{aligned}
        T_{\mathrm{aux}}
        = \mathcal{O}\!\bigl(
        &Ld(Q_C+K_C)
        &+Q_CK_C(d+\log K_C)
        \bigr).
    \end{aligned}
    \label{eq:method_auxiliary_cost}
\end{equation}
At retained token-pair density $\eta$, the selected exact computation costs $\mathcal{O}(\eta L^2d)$.
For fixed cluster counts and a fixed positive $\eta$, the auxiliary cost grows linearly with sequence length and becomes asymptotically smaller than the exact-attention term.
Clustering and token reordering are excluded from this post-clustering complexity.
The supplementary material provides a stage-wise latency breakdown, including clustering, token reordering, output estimation, block scoring, and selection, and examines how these overheads scale with resolution.
\section{Evaluation}
\label{sec:eval}

\begin{table*}
\caption{Quantitative evaluation of our method, \proj, against state-of-the-art sparse attention methods~\cite{xi2025sparse, yang2026sparse, luo2026svoo, zhou2026svg} on five widely adopted vDiTs: HunyuanVideo~\cite{kong2024hunyuanvideo}, Wan2.1-T2V-14B~\cite{wan2025wan}, Wan2.2-T2V-A14B~\cite{wan2025wan22}, LingBot-Video-Dense-1.3B~\cite{lingbot2026video}, and LongCat-Video~\cite{longcat2025video}. We annotate the \goldtext{best} and \silvertext{second-best} results among the sparse methods.}
\resizebox{\textwidth}{!}{
\renewcommand*{\arraystretch}{1}
\begin{tabular}{c|c|cccc|ccccc}
\toprule[0.15em]
\multirow{2}{*}{Model}  & \multirow{2}{*}{Methods} & \multicolumn{4}{c|}{\textbf{Quality Metrics}} & \multicolumn{5}{c}{\textbf{Performance Metrics}}  \\
& & \specialcell{PSNR $\uparrow$\\(dB)} & SSIM$\uparrow$ & LPIPS$\downarrow$ & \specialcell{VBench$\uparrow$\\(\%)} & \specialcell{Attn.\\Speedup$\uparrow$} & \specialcell{Latency $\downarrow$\\(s)} & Speedup$\uparrow$ & \specialcell{FLOPs$\downarrow$\\($10^{15}$)} & \specialcell{Memory$\downarrow$\\(GB)} \\
\midrule[0.05em]
\multirow{6}{*}{HunyuanVideo~\cite{kong2024hunyuanvideo}} & \cellcolor{blue!8} Original & \cellcolor{blue!8} - & \cellcolor{blue!8} - & \cellcolor{blue!8} - & \cellcolor{blue!8} 82.70 & \cellcolor{blue!8} - & \cellcolor{blue!8} 1695.50 & \cellcolor{blue!8} - & \cellcolor{blue!8} 603.4 & \cellcolor{blue!8} 51.0 \\
 & \mode{SVG}~\cite{xi2025sparse} & 25.79 & 0.823 & 0.188 & 82.48 & 1.90 & 1129.97 & 1.50 & 373.2 & \goldtext{53.9} \\
 & \mode{SVG2}~\cite{yang2026sparse} & 29.08 & 0.875 & 0.112 & \silvertext{82.76} & 2.46 & 981.30 & 1.73 & 258.5 & 56.3 \\
 & \mode{SVOO}~\cite{luo2026svoo} & \silvertext{30.20} & \silvertext{0.897} & \silvertext{0.090} & 82.68 & 2.49 & 976.76 & \silvertext{1.74} & 255.9 & 56.4 \\
 & \mode{SVG-EAR}~\cite{zhou2026svg} & 27.60 & 0.843 & 0.176 & 81.87 & \silvertext{2.51} & \silvertext{972.50} & \silvertext{1.74} & \silvertext{242.9} & 55.7 \\
 & \mode{\proj} & \goldtext{31.46} & \goldtext{0.909} & \goldtext{0.082} & \goldtext{82.79} & \goldtext{2.70} & \goldtext{929.49} & \goldtext{1.82} & \goldtext{232.5} & \silvertext{55.2} \\
\midrule[0.05em]
\multirow{6}{*}{Wan2.1-T2V-14B~\cite{wan2025wan}} & \cellcolor{blue!8} Original & \cellcolor{blue!8} - & \cellcolor{blue!8} - & \cellcolor{blue!8} - & \cellcolor{blue!8} 83.40 & \cellcolor{blue!8} - & \cellcolor{blue!8} 1833.20 & \cellcolor{blue!8} - & \cellcolor{blue!8} 652.3 & \cellcolor{blue!8} 47.8 \\
 & \mode{SVG}~\cite{xi2025sparse} & 23.23 & 0.761 & 0.129 & \silvertext{83.32} & 2.03 & 1291.28 & 1.42 & 459.4 & \goldtext{47.8} \\
 & \mode{SVG2}~\cite{yang2026sparse} & 23.20 & 0.754 & 0.132 & 83.19 & \silvertext{2.19} & \silvertext{1252.98} & \silvertext{1.46} & 385.9 & 50.2 \\
 & \mode{SVOO}~\cite{luo2026svoo} & \silvertext{24.79} & \silvertext{0.814} & \silvertext{0.099} & 83.26 & 2.15 & 1258.06 & \silvertext{1.46} & 385.9 & 50.1 \\
 & \mode{SVG-EAR}~\cite{zhou2026svg} & 23.98 & 0.775 & 0.115 & 83.21 & 2.00 & 1294.81 & 1.42 & \silvertext{383.1} & 50.2 \\
 & \mode{\proj} & \goldtext{26.34} & \goldtext{0.844} & \goldtext{0.078} & \goldtext{83.44} & \goldtext{2.37} & \goldtext{1207.77} & \goldtext{1.52} & \goldtext{355.0} & \silvertext{49.9} \\
\midrule[0.05em]
\multirow{6}{*}{Wan2.2-T2V-A14B~\cite{wan2025wan22}} & \cellcolor{blue!8} Original & \cellcolor{blue!8} - & \cellcolor{blue!8} - & \cellcolor{blue!8} - & \cellcolor{blue!8} 81.56 & \cellcolor{blue!8} - & \cellcolor{blue!8} 1341.07 & \cellcolor{blue!8} - & \cellcolor{blue!8} 521.9 & \cellcolor{blue!8} 74.4 \\
 & \mode{SVG}~\cite{xi2025sparse} & 21.26 & 0.740 & 0.140 & 81.51 & 1.98 & 929.74 & \silvertext{1.44} & 376.2 & \goldtext{74.4} \\
 & \mode{SVG2}~\cite{yang2026sparse} & 21.57 & 0.746 & 0.136 & 81.34 & \silvertext{2.13} & \silvertext{928.76} & \silvertext{1.44} & 322.6 & 77.7 \\
 & \mode{SVOO}~\cite{luo2026svoo} & \silvertext{23.20} & \silvertext{0.811} & \silvertext{0.105} & 81.49 & 2.09 & 948.91 & 1.41 & 323.3 & 77.7 \\
 & \mode{SVG-EAR}~\cite{zhou2026svg} & 22.22 & 0.765 & 0.123 & \silvertext{81.52} & 1.96 & 963.68 & 1.39 & \silvertext{320.7} & 77.7 \\
 & \mode{\proj} & \goldtext{23.84} & \goldtext{0.822} & \goldtext{0.097} & \goldtext{81.60} & \goldtext{2.46} & \goldtext{864.11} & \goldtext{1.55} & \goldtext{292.3} & \silvertext{77.4} \\
\midrule[0.05em]
\multirow{6}{*}{LingBot-Video-Dense-1.3B~\cite{lingbot2026video}} & \cellcolor{blue!8} Original & \cellcolor{blue!8} - & \cellcolor{blue!8} - & \cellcolor{blue!8} - & \cellcolor{blue!8} 82.49 & \cellcolor{blue!8} - & \cellcolor{blue!8} 769.45 & \cellcolor{blue!8} - & \cellcolor{blue!8} 230.3 & \cellcolor{blue!8} 32.7 \\
 & \mode{SVG}~\cite{xi2025sparse} & 31.22 & 0.906 & 0.082 & \silvertext{82.27} & 2.64 & 487.42 & 1.58 & 128.7 & \goldtext{32.7} \\
 & \mode{SVG2}~\cite{yang2026sparse} & 31.74 & 0.911 & 0.077 & 82.19 & 2.57 & 490.52 & 1.57 & 114.2 & \silvertext{33.3} \\
 & \mode{SVOO}~\cite{luo2026svoo} & \silvertext{34.76} & \silvertext{0.936} & \silvertext{0.054} & 82.21 & \silvertext{2.65} & \silvertext{477.64} & \silvertext{1.61} & \silvertext{112.0} & \silvertext{33.3} \\
 & \mode{SVG-EAR}~\cite{zhou2026svg} & 32.46 & 0.919 & 0.068 & 82.08 & 2.34 & 502.41 & 1.53 & 113.8 & \silvertext{33.3} \\
 & \mode{\proj} & \goldtext{35.82} & \goldtext{0.941} & \goldtext{0.048} & \goldtext{82.45} & \goldtext{2.79} & \goldtext{464.76} & \goldtext{1.66} & \goldtext{102.5} & \silvertext{33.3} \\
\midrule[0.05em]
\multirow{6}{*}{LongCat-Video~\cite{longcat2025video}} & \cellcolor{blue!8} Original & \cellcolor{blue!8} - & \cellcolor{blue!8} - & \cellcolor{blue!8} - & \cellcolor{blue!8} 84.62 & \cellcolor{blue!8} - & \cellcolor{blue!8} 2346.32 & \cellcolor{blue!8} - & \cellcolor{blue!8} 786.4 & \cellcolor{blue!8} 58.8 \\
 & \mode{SVG}~\cite{xi2025sparse} & 19.85 & 0.690 & 0.193 & 84.66 & \silvertext{3.19} & \goldtext{1408.47} & \goldtext{1.67} & 473.5 & \goldtext{58.8} \\
 & \mode{SVG2}~\cite{yang2026sparse} & 21.59 & 0.748 & 0.149 & \goldtext{84.82} & \goldtext{3.30} & \silvertext{1411.56} & \silvertext{1.66} & \silvertext{396.0} & 62.5 \\
 & \mode{SVOO}~\cite{luo2026svoo} & \silvertext{22.93} & \silvertext{0.790} & \silvertext{0.119} & 84.62 & 3.13 & 1417.71 & \silvertext{1.66} & 397.2 & 62.5 \\
 & \mode{SVG-EAR}~\cite{zhou2026svg} & 22.60 & 0.775 & 0.125 & \silvertext{84.81} & 3.02 & 1443.13 & 1.63 & \silvertext{396.0} & 62.5 \\
 & \mode{\proj} & \goldtext{23.83} & \goldtext{0.817} & \goldtext{0.101} & 84.80 & 3.15 & 1415.46 & \silvertext{1.66} & \goldtext{380.7} & \silvertext{61.7} \\
\bottomrule[0.05em]
\end{tabular}
}
\label{tab:overall}
\end{table*}

\subsection{Experimental Setup}
\label{sec:eva:exp}

\paragraph{Baselines.} 
We evaluate our technique on five video diffusion models: HunyuanVideo-T2V~\cite{kong2024hunyuanvideo}, Wan2.1-T2V-14B~\cite{wan2025wan}, Wan2.2-T2V-A14B~\cite{wan2025wan22}, LingBot-Video-Dense-1.3B~\cite{lingbot2026video} and LongCat-Video~\cite{longcat2025video}.
For HunyuanVideo-T2V, we generate 129-frame videos with a resolution of $720 \times 1280$.
For Wan2.1-T2V-14B, we generate 81-frame videos with a resolution of $720 \times 1280$.
For Wan2.2-T2V-A14B, we generate 81-frame videos with a resolution of $720 \times 1280$.
For LongCat-Video, we generate 93-frame videos with a resolution of $720 \times 1280$. We only run the base text-to-video stage, without the distillation or refinement stages, and disable its built-in block sparse attention.
For LingBot-Video-Dense-1.3B, we generate 121-frame videos with a resolution of $736 \times 1280$.

Our evaluation compares with four sparse attention methods for video diffusion: \mode{SVG}~\cite{xi2025sparse}, \mode{SVG2}~\cite{yang2026sparse}, \mode{SVOO}~\cite{luo2026svoo} and \mode{SVG-EAR}~\cite{zhou2026svg}.

\paragraph{Metrics and Hardware.}
In line with previous studies~\cite{zhao2025pab, zou2025toca}, we adopt VBench~\cite{huang2024vbench} to measure video quality.
We evaluate the generated videos across the 16 VBench dimensions and report the total score computed using the official normalization and weighting scheme.
To assess image quality, we use PSNR, SSIM~\cite{wang2004ssim}, and LPIPS~\cite{zhang2018lpips}, comparing the videos produced by each acceleration method frame by frame with those of the original model.
For efficiency, we report the end-to-end generation latency and speedup, the attention speedup over dense attention, the GPU memory footprint, and the computational cost in FLOPs.
All efficiency results are measured on a single NVIDIA H100 (80 GB) GPU.

\subsection{Accuracy and Performance}
\label{sec:eva:perf}

\paragraph{Configurations.} 
\Tbl{tab:overall} compares \mode{\proj} with state-of-the-art methods. 
\mode{SVG} retains full computation in the first 10 steps, and sparsifies the rest to 45\% density on Wan2.1 and Wan2.2 while 30\% on other models.
\mode{SVG2} retains full computation in the first 10 steps, and sparsifies the rest with top-$p$ = 0.9, reaching a sparsity of 70.4--82.8\%.
\mode{SVOO} retains full computation in the first 10 steps, and sparsifies the rest with top-$p$ = 0.9, reaching a sparsity of 70.1--82.6\%.
\mode{SVG-EAR} retains full computation in the first 10 steps, and sparsifies the rest with top-$p$ = 0.9, reaching a sparsity of 71.0--82.9\%.
\mode{\proj} retains full computation in the first 10 steps, and sparsifies the rest with top-$p$ = 0.8, reaching a sparsity of 79.4--86.0\%.
\proj uses a more aggressive top-$p$ (0.8 vs.\ 0.9) because its additional scoring pass adds overhead, and this setting still gives the best fidelity.
For a consistent comparison, we use a unified sparsity-budgeting scheme for all clustering-based methods, with the same top-p criterion and clustering configuration. In particular, we replace SVOO's original offline layer-wise sparsity profiling with online top-p budgeting, while preserving its co-clustering and block-selection mechanisms. Detailed configurations are provided in the supplementary material.

\paragraph{Video Quality.}
\mode{\proj} achieves the best VBench total score among sparse methods on four of the five models, and is within 0.02 points of the best sparse method on LongCat-Video.
On HunyuanVideo, Wan2.1, and Wan2.2, it achieves the highest total score among all methods, with totals of 82.79\%, 83.44\%, and 81.60\%, all within 0.1 points of the original models.
These results indicate comparable VBench scores to the dense models.
On LingBot, \mode{\proj} (82.45\%) stays closest to the original model (82.49\%) and leads the best sparse baseline (\mode{SVG}, 82.27\%) by 0.18 points.
On LongCat, all sparse methods, including \mode{\proj} (84.80\%), score on par with the original model (84.62\%).
Per-dimension VBench results are provided in the supplementary material.
\begin{figure*}[t]
    \centering
    \includegraphics[width=\textwidth]{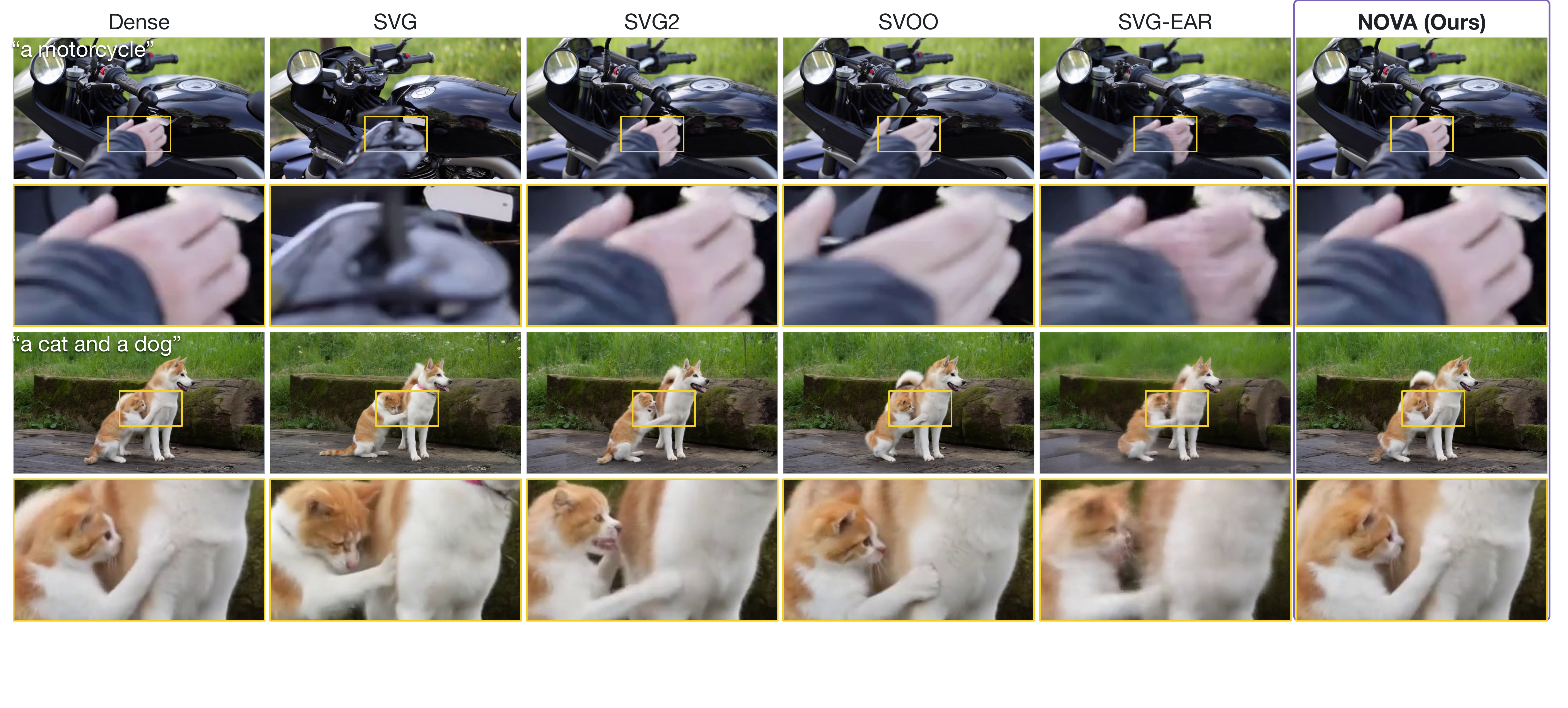}
    \caption{\textbf{Qualitative comparison on HunyuanVideo~\cite{kong2024hunyuanvideo}.} The second row of each example zooms into the yellow box. \mode{\proj} stays closest to dense attention and preserves fine details such as the hand and the cat's face, while other sparse attention methods blur or distort them.}
    \label{fig:examples}
\end{figure*}

\begin{table*}[t]
\centering
\caption{Quality--speed trade-off on HunyuanVideo at three attention densities (720p, 129 frames, 50 steps; quality vs.\ Dense, end-to-end speedup). Density is approximate and measured over the sparse denoising steps. $^\dagger$\Tbl{tab:overall} setting.}
\resizebox{\textwidth}{!}{
\renewcommand*{\tabcolsep}{3pt}
\begin{tabular}{c|cccc|cccc|cccc|cccc|cccc}
\toprule[0.15em]
\multirow{2}{*}{Reference Density} & \multicolumn{4}{c|}{\mode{SVG}} & \multicolumn{4}{c|}{\mode{SVG2}} & \multicolumn{4}{c|}{\mode{SVOO}} & \multicolumn{4}{c|}{\mode{SVG-EAR}} & \multicolumn{4}{c}{\proj} \\
 & PSNR$\uparrow$ & SSIM$\uparrow$ & LPIPS$\downarrow$ & Spd.$\uparrow$ & PSNR$\uparrow$ & SSIM$\uparrow$ & LPIPS$\downarrow$ & Spd.$\uparrow$ & PSNR$\uparrow$ & SSIM$\uparrow$ & LPIPS$\downarrow$ & Spd.$\uparrow$ & PSNR$\uparrow$ & SSIM$\uparrow$ & LPIPS$\downarrow$ & Spd.$\uparrow$ & PSNR$\uparrow$ & SSIM$\uparrow$ & LPIPS$\downarrow$ & Spd.$\uparrow$ \\
\midrule[0.05em]
$\sim$20\% & 21.89 & 0.733 & 0.345 & 1.60$\times$ & 26.18 & 0.846 & 0.151 & 1.85$\times$ & 27.56 & 0.862 & 0.124 & 1.84$\times$ & 26.02 & 0.804 & 0.311 & 1.88$\times$ & \best{29.02} & \best{0.891} & \best{0.087} & 1.89$\times$ \\
$\sim$22\% & 22.56 & 0.744 & 0.303 & 1.51$\times$ & 26.23 & 0.849 & 0.144 & 1.79$\times$ & 28.06 & 0.872 & 0.113 & 1.78$\times$ & 26.19 & 0.809 & 0.295 & 1.80$\times$ & \best{29.05} & \best{0.891} & \best{0.088} & 1.84$\times$ \\
$\sim$25\%$^\dagger$ & 25.79 & 0.823 & 0.188 & 1.50$\times$ & 29.08 & 0.875 & 0.112 & 1.73$\times$ & 30.20 & 0.897 & 0.090 & 1.74$\times$ & 27.60 & 0.843 & 0.176 & 1.74$\times$ & \best{31.46} & \best{0.909} & \best{0.082} & 1.82$\times$ \\
\bottomrule[0.15em]
\end{tabular}
}
\label{tab:sensitivity}
\end{table*}

\paragraph{Image Quality.}
Across PSNR, SSIM, and LPIPS, \mode{\proj} outperforms prior methods on all five models.
On HunyuanVideo, \mode{\proj} achieves 31.46~dB PSNR vs. 30.20~dB from the best baseline (\mode{SVOO}), together with the best SSIM (0.909) and LPIPS (0.082).
Similar gains are observed on the other models, e.g., 26.34~dB vs. 24.79~dB on Wan2.1 and 35.82~dB vs. 34.76~dB on LingBot, indicating strong generalization.
\Fig{fig:examples} illustrates qualitative results, showing \mode{\proj} retains finer details compared to other methods.
Additional qualitative comparisons across all five models are provided in the supplementary material.

\paragraph{Performance.}
\mode{\proj} achieves the highest end-to-end speedup on HunyuanVideo, Wan2.1, Wan2.2, and LingBot, reaching $1.82\times$, $1.52\times$, $1.55\times$, and $1.66\times$.
On LongCat, it reaches $1.66\times$, within $0.01\times$ of the fastest baseline.

\paragraph{Computation.}
We further compare the computation over the entire generation.
\mode{\proj} requires the fewest FLOPs on all five models, reducing the total by 44.0\% to 61.5\% relative to the original model (\eg, 652.3 to 355.0~PFLOPs on Wan2.1 and 603.4 to 232.5~PFLOPs on HunyuanVideo) and by 3.9\% to 8.9\% relative to the lowest-FLOP baseline.

\subsection{Ablation Study}

\begin{table}[t]
\centering
\caption{Ablation on Wan2.1-T2V-14B (720p, 81 frames, 50 steps; quality vs.\ Dense). All variants use centroid compensation and retain the same 25\% of attention.}
\resizebox{\columnwidth}{!}{
\renewcommand*{\arraystretch}{1}
\renewcommand*{\tabcolsep}{3pt}
\begin{tabular}{c|cc|ccc}
\toprule[0.15em]
Variant & Clustering & Block score & PSNR$\uparrow$ & SSIM$\uparrow$ & LPIPS$\downarrow$ \\
\midrule[0.05em]
\mode{SVG-EAR} & k-means & Unnormalized error & 21.05 & 0.718 & 0.169 \\
\mode{SVG2} + our score & k-means & Normalized error & 26.20 & 0.840 & 0.080 \\
\mode{SVOO-EAR} & Co-clustering & Unnormalized error & 24.16 & 0.809 & 0.102 \\
\proj & Co-clustering & Normalized error & \textbf{26.34} & \textbf{0.844} & \textbf{0.078} \\
\bottomrule[0.15em]
\end{tabular}
}
\label{tab:ablation}
\end{table}

\Tbl{tab:ablation} shows the contributions of individual components in \mode{\proj} via an ablation study on Wan2.1.
We vary the two components: the clustering (independent query/key k-means as in \mode{SVG2}, or QK-profile co-clustering) and the block score (the residual score of \mode{SVG-EAR}, or our denominator-aware score).
All four variants use centroid compensation and retain the same 25\% of attention, so differences come only from which blocks are kept.

The denominator-aware score is the larger contributor.
With \mode{SVG2} clustering, it raises PSNR from 21.05~dB to 26.20~dB and lowers LPIPS from 0.169 to 0.080 at the same budget.
With co-clustering, it improves PSNR from 24.16~dB (\mode{SVOO-EAR}) to 26.34~dB.
Co-clustering is also beneficial: with the residual score, it adds 3.11~dB over k-means, and with our score, it adds a further 0.14~dB.
Combining both, \mode{\proj} attains the highest PSNR, SSIM, and LPIPS among all variants.

\subsection{Sensitivity Study}
\label{sec:eval:sens}

\Tbl{tab:sensitivity} compares the quality--speed trade-offs of all methods on HunyuanVideo at three operating points.
The row labels indicate the approximate attention densities.
At every density, \mode{\proj} achieves the best PSNR, SSIM, and LPIPS while running the fastest, \eg, 29.02~dB at 1.89$\times$ versus 27.56~dB at 1.84$\times$ for the best baseline, \mode{SVOO}, at 20\% density.
In contrast, \mode{SVG} loses most of its speedup as its density grows, dropping from 1.60$\times$ to 1.51$\times$, and \mode{SVG2} and \mode{SVG-EAR} gain less than 0.2~dB when the density increases from 20\% to 22\%.
Thus, \mode{\proj} does not depend on a particular setting.

\subsection{Scalability Study}
\label{sec:eval:scal}

\begin{figure}[t]
\centering
\includegraphics[width=\columnwidth]{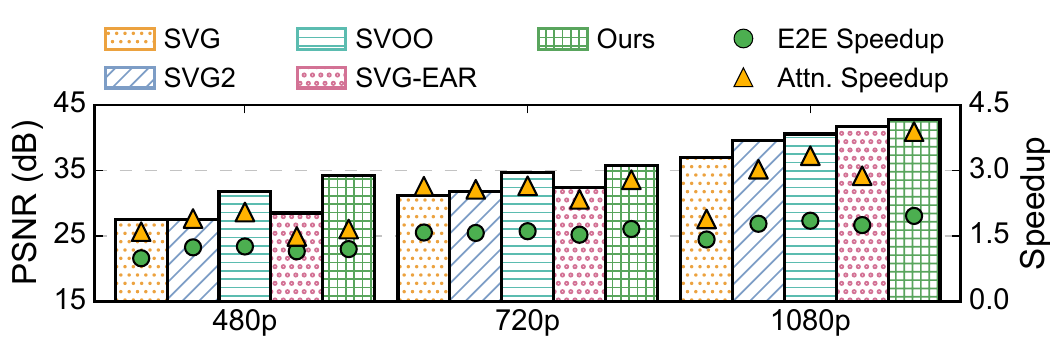}
\caption{Quality and speedup of all methods on LingBot at three resolutions. Bars show PSNR and markers show end-to-end and attention speedup over the original model.}
\label{fig:scalability}
\end{figure}

\Fig{fig:scalability} compares all methods on LingBot at 480p, 720p, and 1080p using their main-table settings.
\mode{\proj} achieves the highest PSNR at every resolution.
At 480p, its clustering and scoring overhead limits end-to-end speedup to $1.20\times$, below \mode{SVOO} ($1.26\times$).
As resolution increases, this overhead is better amortized: at 1080p, \mode{\proj} reaches $1.96\times$ end-to-end and $3.89\times$ attention speedup, versus $1.85\times$ and $3.34\times$ for \mode{SVOO}.
Dense attention grows quadratically with the number of tokens while the overhead of \mode{\proj} grows linearly, so \mode{\proj} scales favorably to the higher resolutions of recent video generators.
\section{Conclusion}
\label{sec:conc}
We presented \proj, a sparse attention method for video diffusion transformers that selects blocks by their effect on the normalized attention output rather than by attention mass alone.
\proj first estimates the output from size-weighted centroids, then uses this estimate to score blocks while accounting for both value approximation and softmax renormalization.
On five video models, \proj achieves the best PSNR, SSIM, and LPIPS among sparse attention methods and a 1.52$\times$ to 1.82$\times$ end-to-end speedup.
\section*{LLM Usage Statement}
We used Claude (Anthropic) as an LLM assistant in this work, and the authors take full responsibility for all submitted content.
\textbf{Related work and citations:} the authors used the LLM to assist with literature searches and identify potentially relevant references; the authors wrote the related work section and verified every citation against the original source.
\textbf{Writing:} the LLM was used to edit and refine author-written text (language, grammar, and restructuring).
\textbf{Research ideas, code, experiments, and figures:} the LLM was used only in an assistive role to refine author-produced content; all ideas, claims, implementations, experimental designs, results, and figures originate from the authors.
\textbf{Data:} no LLM or AI tool was used for data creation or annotation.
\textbf{Feedback:} the full manuscript was given to the LLM to obtain reviewer-style feedback, which the authors considered when revising the paper.

{\small
\bibliographystyle{ieeenat_fullname}
\bibliography{ref}
}
\clearpage
\maketitlesupplementary

\appendix

\newtheorem{lemma}{Lemma}
\newtheorem{proposition}{Proposition}
\newenvironment{proofs}{\noindent\textit{Proof.}\ }{\hfill$\square$\par\smallskip}

This supplementary material contains:
\begin{itemize}[leftmargin=*,itemsep=0pt,topsep=2pt]
    \item \Sect{sec:supp_theory}: derivations of the joint output error and the bound underlying block selection, together with empirical validation of the first-order approximation.
    \item \Sect{sec:supp_overhead}: latency breakdowns across models and resolutions, including clustering, scoring, and attention computation.
    \item \Sect{sec:supp_vbench}: per-dimension VBench results and details of score aggregation.
    \item \Sect{sec:supp_hyperparameters}: hyperparameters for the main evaluation and the HunyuanVideo sensitivity study.
    \item \Sect{sec:supp_qual}: additional qualitative comparisons across all five video models.
\end{itemize}
\section{Theoretical Analysis}
\label{sec:supp_theory}

We extend the normalized output-error analysis in \Sect{sec:analysis} to jointly account for all unselected blocks and establish the basis of the block score in Eq.~\eqref{eq:method_block_score}.
We further empirically validate the accuracy of the first-order approximation in Eq.~\eqref{eq:method_output_change} across five video diffusion transformers.

\paragraph{Joint error of the unselected blocks.}
Fix a query cluster $\mathcal Q_i$ and evaluate the residuals at its centroid $\bar q_i$, as in the main paper.
Let $\mathcal J_i$ denote the selected key clusters and $\mathcal D_i=\{1,\ldots,K_C\}\setminus\mathcal J_i$ the unselected ones.
Refining all selected blocks at once changes the coarse numerator and denominator by $\sum_{j\in\mathcal J_i}\Delta N_{ij}$ and $\sum_{j\in\mathcal J_i}\Delta Z_{ij}$.
Since $\hat N_i=\hat O_i\hat Z_i$, the same algebra as in Eq.~\eqref{eq:method_output_change} gives the exact change of the output,
\begin{equation}
    \tilde O_i-\hat O_i
    =
    \frac{\sum_{j\in\mathcal J_i}\bigl(\Delta N_{ij}-\hat O_i\Delta Z_{ij}\bigr)}
         {\hat Z_i+\sum_{j\in\mathcal J_i}\Delta Z_{ij}},
    \label{eq:supp_exact_change}
\end{equation}
where $\tilde O_i$ is the refined output of Eq.~\eqref{eq:method_refined_output} evaluated at $\bar q_i$.
Thus the numerator is additive over blocks and every block enters through its centered residual of Eq.~\eqref{eq:method_centered_residual}.
If every block were refined, the output would be exact.
The error that remains after refining $\mathcal J_i$ is therefore the contribution of the unselected blocks, which to first order is
\begin{equation}
    \Delta O_i^{(1)}
    =
    \frac{1}{\hat Z_i}
    \sum_{j\in\mathcal D_i}
    \sum_{n\in\mathcal K_j}r_{ijn}.
    \label{eq:supp_joint_error}
\end{equation}
Directly minimizing its squared norm couples the block decisions through inner products between their residuals.
The benefit of selecting one block depends on which other blocks are selected, rather than on that block alone.
Exact optimization therefore requires a combinatorial search over feasible subsets, while a sequential selection procedure must update candidate benefits as the selected set changes.
Such dependencies are costly in the attention execution path, where block scores should be computed in parallel.
The additive residual energy used by \proj provides an upper bound that can be evaluated independently for each block.

\begin{proposition}
\label{prop:supp_bound}
Let $M_i=\sum_{j\in\mathcal D_i}n_j^K$ be the number of key tokens in the unselected blocks of query cluster $i$.
Then
\begin{equation}
    \left\|\Delta O_i^{(1)}\right\|_2^2
    \le
    \frac{M_i}{\hat Z_i^2}
    \sum_{j\in\mathcal D_i}
    \sum_{n\in\mathcal K_j}\|r_{ijn}\|_2^2.
    \label{eq:supp_error_bound}
\end{equation}
\end{proposition}

\begin{proofs}
The sum in Eq.~\eqref{eq:supp_joint_error} contains $M_i$ residual vectors.
Applying the triangle inequality and then the Cauchy--Schwarz inequality gives
\begin{equation}
\begin{aligned}
    \left\|
    \sum_{j\in\mathcal D_i}
    \sum_{n\in\mathcal K_j}r_{ijn}
    \right\|_2^2
    &\le
    \left(
    \sum_{j\in\mathcal D_i}
    \sum_{n\in\mathcal K_j}\|r_{ijn}\|_2
    \right)^2\\
    &\le
    M_i
    \sum_{j\in\mathcal D_i}
    \sum_{n\in\mathcal K_j}\|r_{ijn}\|_2^2.
\end{aligned}
\end{equation}
Dividing by $\hat Z_i^2$ proves Eq.~\eqref{eq:supp_error_bound}.
\end{proofs}

\paragraph{Score as the minimizer of the bound.}
For a fixed retained token-pair area, $M_i$ is fixed, and $\hat Z_i$ is independent of the selection.
Reducing the residual energy of the unselected blocks therefore reduces this upper bound.
Write $E_{ij}=\sum_{n\in\mathcal K_j}\|r_{ijn}\|_2^2$ for the residual energy of block $B_{ij}$.
Minimizing the bound in Eq.~\eqref{eq:supp_error_bound} under a token-pair budget $R_i$ is equivalent to
\begin{equation}
    \max_{\mathcal J_i}\ \sum_{j\in\mathcal J_i}E_{ij}
    \quad\text{s.t.}\quad
    \sum_{j\in\mathcal J_i}C_{ij}\le R_i,
    \label{eq:supp_knapsack}
\end{equation}
because $\sum_{j\in\mathcal D_i}E_{ij}=\sum_{j}E_{ij}-\sum_{j\in\mathcal J_i}E_{ij}$ and the first term does not depend on the selection.
Refining block $B_{ij}$ thus removes $E_{ij}$ from the objective at cost $C_{ij}=n_i^Qn_j^K$, and the ratio of the two is precisely $s_{ij}$ in Eq.~\eqref{eq:method_block_score}.

\begin{proposition}
\label{prop:supp_greedy}
Sort the blocks of query cluster $i$ by $s_{ij}$ in descending order and select them until the budget $R_i$ is reached.
The resulting selection solves the linear relaxation of Eq.~\eqref{eq:supp_knapsack}, in which blocks may be refined fractionally, and differs from it in at most one block, namely the last block that is only partially covered by the budget.
\end{proposition}

\begin{proofs}
Eq.~\eqref{eq:supp_knapsack} is a 0--1 knapsack problem with values $E_{ij}$ and weights $C_{ij}$.
Its linear relaxation is the fractional knapsack problem, whose optimum is obtained by filling the budget in descending order of the value-to-weight ratio $E_{ij}/C_{ij}=s_{ij}$ and taking a fraction of the first block that no longer fits.
Selecting blocks in descending order of $s_{ij}$ reproduces this solution for every block before the boundary block.
\end{proofs}

The score therefore prioritizes residual-energy reduction per unit computation cost.
Together with Proposition~\ref{prop:supp_bound}, this allows \proj to control the joint first-order error bound through block scores that are computed independently and in parallel, without any sequential update across blocks.

\begin{figure}[t]
    \centering
    \includegraphics[width=\columnwidth]{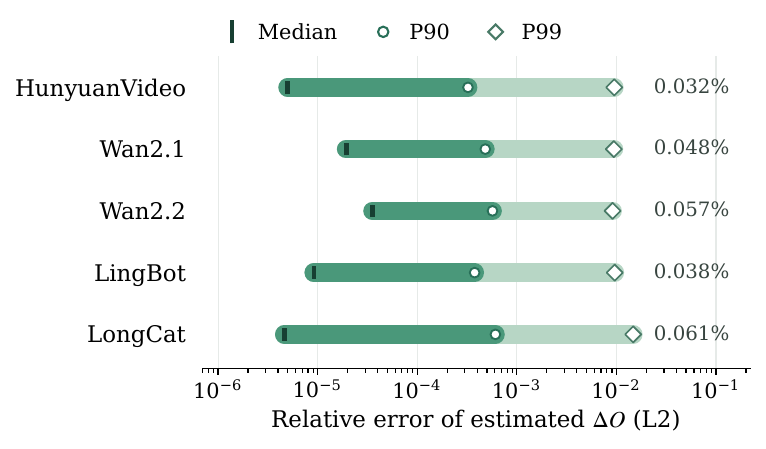}
    \caption{\textbf{Accuracy of the first-order output-change approximation.}
    Distribution of the relative $\ell_2$ error between the first-order approximation in Eq.~\eqref{eq:method_output_change} and the exact single-block output change across five video diffusion transformers.
    Markers and intervals indicate the median, 90th, and 99th percentiles.}
    \label{fig:supp_delta_o}
\end{figure}

\paragraph{Accuracy of the first-order approximation.}
Eq.~\eqref{eq:method_output_change} approximates the output change caused by refining an individual block by replacing the perturbed denominator $\hat Z_i+\Delta Z_{ij}$ with $\hat Z_i$.
To evaluate the accuracy of this approximation, we compare the estimated output change against the exact change computed by replacing each block's centroid contribution with token-level attention, while keeping all other blocks approximated by centroids.
We measure the relative $\ell_2$ error across five video diffusion transformers.
\Fig{fig:supp_delta_o} presents the error distributions, showing that the first-order approximation remains accurate across different models.

\section{Latency Breakdown of \proj}
\label{sec:supp_overhead}

\paragraph{Setting.}
We profile the attention of one sparse denoising step of \proj on five models at 720p-class: LingBot-Video-Dense-1.3B (121 frames, 40 steps), Wan2.1-T2V-14B (81 frames, 50 steps), Wan2.2-T2V-A14B (81 frames, 40 steps), LongCat-Video (93 frames, 50 steps), and HunyuanVideo (129 frames, 50 steps).
We also profile LingBot-Video-Dense-1.3B at the three resolutions of \Fig{fig:scalability} ($480\times832$, $736\times1280$, and $1088\times1920$).
All runs use the same \proj method and stage-timing protocol as the main evaluation, with the top-$p$ budget, centroid counts, clustering iterations, and dense first layer taken from the same profiling configuration.
Each stage is bracketed by device synchronization and timed with CUDA events, and the times are averaged over denoising steps 20, 25, and 30 and 5 prompts.
All measurements were taken on a single NVIDIA H100(80GB) GPU.

\begin{figure}[t]
    \centering
    \includegraphics[width=\columnwidth]{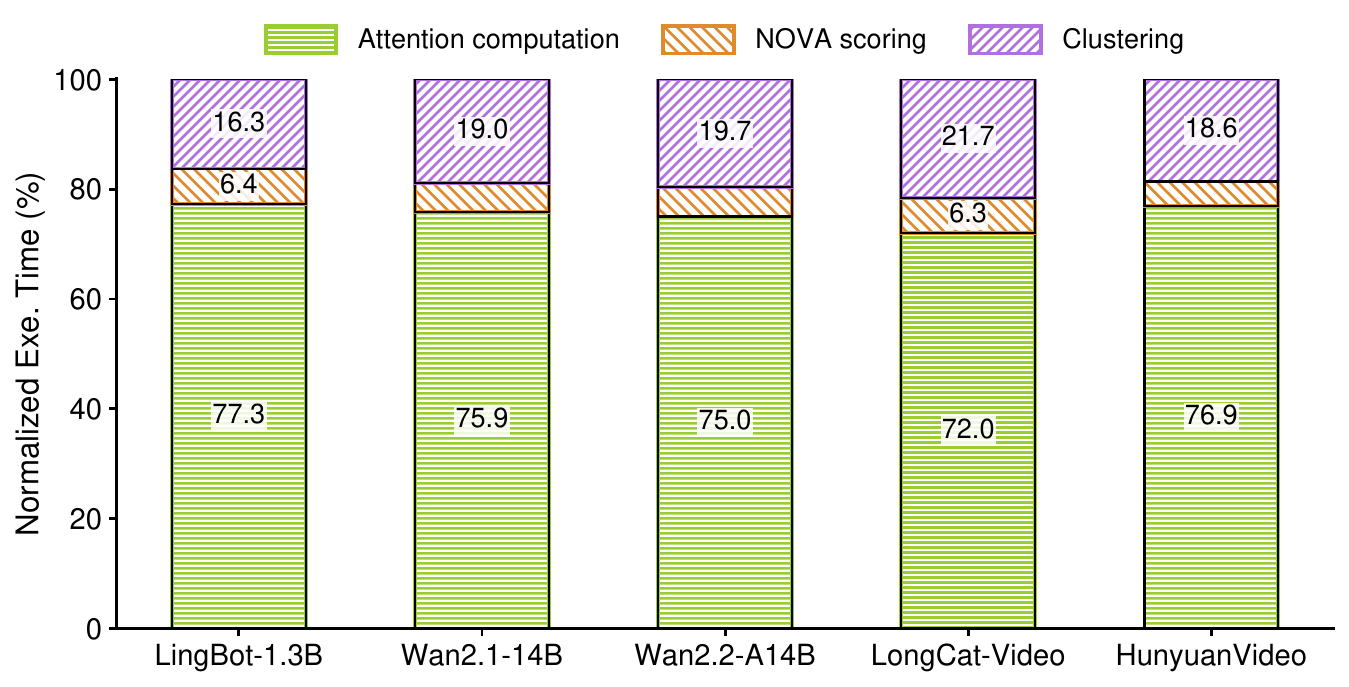}
    \caption{\textbf{Breakdown of attention time across models at 720p-class resolutions.} Each bar is one sparse denoising step, normalized to its total attention time (\%).}
    \label{fig:supp_overhead_model}
\end{figure}

\begin{figure}[t]
    \centering
    \includegraphics[width=\columnwidth]{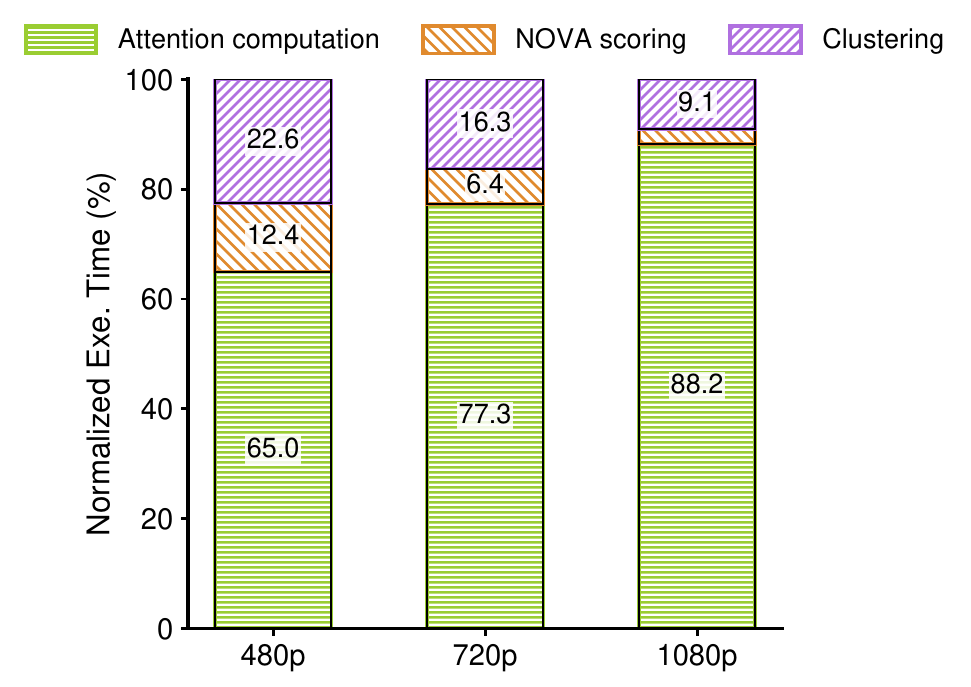}
    \caption{Breakdown of attention time across resolutions for LingBot-Video-Dense-1.3B.}
    \label{fig:supp_overhead_res}
\end{figure}

\paragraph{Groups.}
\Fig{fig:supp_overhead_model} and \Fig{fig:supp_overhead_res} divide the attention time of a step into three groups, each shown as a share of the total attention time of that step:
\emph{clustering} covers QK co-clustering with warm-started centroids, the reordering of queries, keys, and values into contiguous blocks (and its inverse), and the value centroids;
\emph{NOVA scoring} covers the coarse output of Eq.~\eqref{eq:method_coarse_output} with the fused block-scoring kernel of Eq.~\eqref{eq:method_block_score}, and the top-$p$ budget with the selection of the highest-scoring blocks;
\emph{attention computation} covers the block-sparse attention with FlashInfer~\cite{ye2025flashinfer} (including planning), the Triton kernel that merges the exact output with the centroid terms under the shared normalization of Eq.~\eqref{eq:method_refined_output}, untracked glue, and the dense first layer.

\paragraph{Clustering and scoring are a small and shrinking share.}
Across the five models at 720p-class resolutions, clustering and NOVA scoring together take between 23\% and 28\% of the attention time, and attention computation takes between 72\% and 77\% (\Fig{fig:supp_overhead_model}).
The clustering share is largest for LongCat-Video (21.7\%) and smallest for LingBot (16.3\%), while the scoring share stays between 4.5\% and 6.4\% for all models.
For LingBot, the share of clustering and scoring falls from 35.0\% at 480p to 22.7\% at 720p and 11.9\% at 1080p (\Fig{fig:supp_overhead_res}).
In absolute terms, clustering and scoring grow by about $4\times$ from 480p to 1080p, while the sparse attention itself (plan and run) grows by about $20\times$, so the share of \proj-specific work keeps falling as the number of tokens increases.

\section{Full VBench Results}
\label{sec:supp_vbench}

VBench~\cite{huang2024vbench} is a benchmark for text-to-video generation that decomposes video quality into 16 dimensions, each evaluated automatically on the generated videos.
The dimensions are grouped into a \emph{quality score} and a \emph{semantic score}.
The quality score covers subject consistency, background consistency, temporal flickering, motion smoothness, dynamic degree, aesthetic quality, and imaging quality.
The semantic score covers object class, multiple objects, human action, color, spatial relationship, scene, appearance style, temporal style, and overall consistency.
Each dimension is scored with a dedicated evaluator, such as feature-similarity models for consistency, optical flow for motion, and object detectors for object-level dimensions.
Raw scores are normalized to a common range with the official reference bounds and averaged within each group.
The total score is the weighted combination of the two group scores, $\text{Total}=(4\cdot\text{Quality}+\text{Semantic})/5$.

All videos are generated with the settings of \Sect{sec:eva:perf}, using the 946 prompts of the VBench evaluation set and 5 random seeds per prompt.
\Tbl{tab:supp_vbench} reports the per-dimension scores and the two group scores for every model and method in \Tbl{tab:overall}.
The total scores are identical to the VBench column of \Tbl{tab:overall}.
\begin{table}[t]
\centering
\caption{Hyperparameters of the main evaluation (\Tbl{tab:overall}). All methods keep full attention for the first 10 denoising steps and the first 3\% of layers dense. All clustering-based methods use $Q_C{=}256$ query and $K_C{=}1024$ key centroids with two clustering iterations per step, warm-started from the previous step. SVG is set by density; \emph{Others} (SVG2, SVOO, SVG-EAR) and \proj are set by top-$p$. LingBot uses $736\times1280$.}
\label{tab:supp_hyper_main}
\small
\setlength{\tabcolsep}{3.5pt}
\begin{tabular}{@{}lcccccc@{}}
\toprule
& \multicolumn{3}{c}{Generation} & \multicolumn{3}{c}{Sparsity setting} \\
\cmidrule(lr){2-4}\cmidrule(l){5-7}
Model & Res. & Frames & Steps & SVG & Others & \proj \\
\midrule
HunyuanVideo & 720p & 129 & 50 & 0.30 & 0.9 & 0.8 \\
Wan2.1       & 720p & 81  & 50 & 0.45 & 0.9 & 0.8 \\
Wan2.2       & 720p & 81  & 40 & 0.45 & 0.9 & 0.8 \\
LongCat      & 720p & 93  & 50 & 0.30 & 0.9 & 0.8 \\
LingBot      & 736p & 121 & 40 & 0.30 & 0.9 & 0.8 \\
\bottomrule
\end{tabular}
\end{table}

\begin{table}[t]
\centering
\caption{Settings of the density sweep on HunyuanVideo (\Tbl{tab:sensitivity}). SVG uses a fixed density; all other methods use top-$p$.}
\label{tab:supp_hyper_sens}
\small
\setlength{\tabcolsep}{6pt}
\begin{tabular}{@{}lccccc@{}}
\toprule
Density & SVG & SVG2 & SVOO & SVG-EAR & \proj \\
\midrule
$\sim$20\% & 0.20 & 0.80 & 0.80 & 0.80 & 0.60 \\
$\sim$22\% & 0.25 & 0.85 & 0.85 & 0.85 & 0.75 \\
$\sim$25\% & 0.30 & 0.90 & 0.90 & 0.90 & 0.80 \\
\bottomrule
\end{tabular}
\end{table}
\begin{table*}[t]
\centering
\caption{Full VBench results (\%) for all models and methods in \Tbl{tab:overall}. Dimensions are listed in the order of the VBench quality and semantic groups; \emph{Quality}, \emph{Semantic}, and \emph{Total} are the aggregate scores. Total $=(4\cdot$Quality$+$Semantic$)/5$. Bold: best among sparse methods within each model.}
\label{tab:supp_vbench}
\resizebox{\textwidth}{!}{
\renewcommand*{\tabcolsep}{2.5pt}
\begin{tabular}{ll|ccccccc|ccccccccc|ccc}
\toprule[0.15em]
Model & Method & \shortstack{Subj.\\cons.} & \shortstack{Bg.\\cons.} & \shortstack{Temp.\\flick.} & \shortstack{Motion\\smooth.} & \shortstack{Dyn.\\deg.} & \shortstack{Aesth.} & \shortstack{Imag.} & \shortstack{Overall\\cons.} & \shortstack{Obj.\\class} & \shortstack{Mult.\\obj.} & \shortstack{Human\\act.} & \shortstack{Color} & \shortstack{Spatial} & \shortstack{Scene} & \shortstack{Temp.\\style} & \shortstack{App.\\style} & \shortstack{Quality} & \shortstack{Semantic} & \shortstack{Total} \\
\midrule[0.05em]
\multirow{6}{*}{HunyuanVideo} & Dense & 95.70 & 96.89 & 99.75 & 99.44 & 65.66 & 61.42 & 64.77 & 24.26 & 90.63 & 71.64 & 94.40 & 91.43 & 69.40 & 52.27 & 23.54 & 21.08 & 84.31 & 76.24 & 82.70 \\
 & SVG & 95.45 & 96.74 & 99.86 & 99.32 & 62.36 & 62.52 & 63.87 & 24.61 & 85.85 & 71.32 & 95.61 & 90.58 & 67.92 & \textbf{58.93} & 23.08 & 21.11 & 84.00 & 76.43 & 82.48 \\
 & SVG2 & 95.01 & 96.35 & \textbf{99.95} & 99.31 & 66.04 & \textbf{62.63} & 64.96 & 25.98 & \textbf{92.60} & 68.79 & 93.69 & 91.00 & 72.51 & 49.53 & 22.71 & 21.58 & 84.33 & 76.46 & 82.76 \\
 & SVOO & \textbf{97.04} & 96.91 & 99.77 & 99.44 & 63.17 & 61.24 & \textbf{65.17} & 23.17 & \textbf{92.60} & \textbf{73.91} & 95.80 & 87.80 & 67.93 & 52.45 & 22.93 & 20.93 & \textbf{84.41} & 75.75 & 82.68 \\
 & SVG-EAR & 93.93 & \textbf{96.92} & 99.73 & \textbf{99.57} & \textbf{67.61} & 60.14 & 60.91 & \textbf{26.49} & 90.19 & 73.29 & 90.81 & 83.24 & 73.59 & 44.60 & \textbf{24.20} & \textbf{21.81} & 83.42 & 75.66 & 81.87 \\
 & \proj & 96.36 & 96.87 & 99.81 & 99.14 & 67.44 & 60.64 & 64.84 & 22.65 & 89.96 & 72.90 & \textbf{96.89} & \textbf{92.62} & \textbf{74.76} & 52.33 & 23.48 & 20.73 & 84.32 & \textbf{76.67} & \textbf{82.79} \\
\midrule[0.05em]
\multirow{6}{*}{Wan2.1-14B} & Dense & 96.57 & 97.29 & 99.69 & 97.81 & 62.70 & 65.80 & 70.51 & 25.08 & 89.99 & 79.20 & 90.61 & 92.01 & 83.37 & 31.92 & 22.52 & 25.17 & 85.00 & 76.99 & 83.40 \\
 & SVG & 96.80 & 97.46 & 99.52 & 97.95 & 61.11 & 66.04 & 69.69 & 25.14 & \textbf{90.17} & 78.25 & 89.25 & 90.61 & 84.29 & 33.39 & 22.90 & \textbf{25.48} & 84.86 & \textbf{77.15} & 83.32 \\
 & SVG2 & 95.59 & 97.44 & 99.74 & 97.70 & 66.04 & 66.29 & 67.92 & 25.35 & 88.84 & 81.24 & 89.42 & 90.24 & 84.26 & 31.88 & 22.49 & 25.19 & 84.75 & 76.93 & 83.19 \\
 & SVOO & 94.80 & 97.54 & 99.77 & \textbf{98.17} & 61.02 & \textbf{66.43} & \textbf{69.95} & 25.30 & 88.30 & 77.01 & 90.12 & 91.27 & \textbf{85.78} & \textbf{35.35} & 22.30 & 24.60 & 84.84 & 76.93 & 83.26 \\
 & SVG-EAR & \textbf{97.55} & \textbf{98.66} & 99.35 & 97.75 & 60.85 & 66.03 & 68.50 & 24.82 & 82.20 & \textbf{81.71} & \textbf{92.05} & \textbf{94.45} & 83.10 & 30.52 & \textbf{23.29} & 24.69 & 84.87 & 76.58 & 83.21 \\
 & \proj & 97.07 & 97.26 & \textbf{99.99} & 97.97 & \textbf{67.60} & 64.04 & 68.78 & \textbf{26.18} & 89.02 & 73.48 & 91.76 & 88.70 & 83.59 & 34.38 & 22.72 & 24.98 & \textbf{85.13} & 76.68 & \textbf{83.44} \\
\midrule[0.05em]
\multirow{6}{*}{Wan2.2-A14B} & Dense & 95.68 & 94.10 & 99.14 & 96.29 & 74.59 & 64.65 & 70.54 & 24.52 & 88.92 & 65.47 & 80.61 & 92.31 & 86.81 & 27.99 & 20.69 & 22.33 & 83.88 & 72.28 & 81.56 \\
 & SVG & \textbf{96.94} & 94.02 & 99.03 & 96.26 & 77.17 & 65.09 & 67.19 & 23.25 & \textbf{96.96} & 61.84 & 78.98 & 92.31 & 86.87 & 30.88 & 20.62 & 21.82 & 83.79 & \textbf{72.38} & 81.51 \\
 & SVG2 & 94.96 & 94.02 & 99.07 & 96.18 & 75.69 & 64.45 & 70.49 & \textbf{26.07} & 90.37 & 59.42 & 80.16 & \textbf{92.78} & 88.84 & 23.30 & 20.47 & \textbf{22.66} & 83.70 & 71.90 & 81.34 \\
 & SVOO & 96.05 & 94.16 & 98.57 & \textbf{96.48} & 71.31 & 64.96 & 71.55 & 23.24 & 87.96 & \textbf{71.03} & \textbf{80.62} & 92.50 & 87.30 & 28.24 & \textbf{21.00} & 21.68 & 83.78 & 72.35 & 81.49 \\
 & SVG-EAR & 95.97 & 94.41 & \textbf{99.20} & 96.19 & \textbf{77.82} & 63.41 & 69.64 & 23.59 & 88.92 & 62.69 & 80.14 & 90.96 & 88.42 & \textbf{31.01} & 20.57 & 22.36 & 83.89 & 72.04 & 81.52 \\
 & \proj & 93.57 & \textbf{96.05} & 99.04 & 96.11 & 73.58 & \textbf{65.86} & \textbf{71.88} & 25.22 & 84.20 & 62.36 & 79.14 & 88.79 & \textbf{89.98} & 29.16 & 20.79 & 22.47 & \textbf{84.09} & 71.66 & \textbf{81.60} \\
\midrule[0.05em]
\multirow{6}{*}{LingBot-1.3B} & Dense & 96.09 & 95.86 & 99.76 & 99.18 & 70.02 & 62.32 & 62.40 & 24.86 & 91.95 & 78.78 & 91.26 & 91.43 & 66.86 & 48.80 & 22.04 & 21.25 & 84.15 & 75.87 & 82.49 \\
 & SVG & 94.97 & 94.34 & 99.66 & 99.04 & 70.63 & 62.60 & \textbf{63.63} & 25.14 & \textbf{95.57} & 81.49 & 88.65 & 91.36 & \textbf{70.17} & 47.52 & 21.84 & 20.55 & 83.79 & 76.22 & 82.27 \\
 & SVG2 & \textbf{97.16} & 94.55 & 99.45 & 98.83 & 64.99 & 63.41 & 63.09 & 23.83 & 90.66 & 81.43 & 91.29 & 89.72 & 67.30 & \textbf{54.94} & \textbf{22.50} & 20.89 & 83.63 & \textbf{76.40} & 82.19 \\
 & SVOO & 96.31 & 94.42 & 99.56 & \textbf{99.31} & \textbf{76.36} & 61.89 & 58.56 & 24.22 & 91.25 & 78.46 & 91.28 & \textbf{93.56} & 69.28 & 50.55 & 22.21 & 20.96 & 83.70 & 76.24 & 82.21 \\
 & SVG-EAR & 96.28 & 94.20 & 99.76 & 98.69 & 68.00 & \textbf{63.92} & 61.59 & \textbf{25.95} & 95.04 & \textbf{83.07} & \textbf{91.59} & 89.47 & 68.11 & 44.55 & 22.17 & 20.76 & 83.54 & 76.25 & 82.08 \\
 & \proj & 95.12 & \textbf{95.75} & \textbf{99.93} & \textbf{99.31} & 68.88 & 63.00 & 62.62 & 24.05 & 92.98 & 80.49 & 90.67 & 92.50 & 64.17 & 49.62 & 21.88 & \textbf{21.12} & \textbf{84.14} & 75.69 & \textbf{82.45} \\
\midrule[0.05em]
\multirow{6}{*}{LongCat-Video} & Dense & 96.79 & 97.42 & 98.95 & 98.54 & 78.02 & 68.50 & 71.83 & 24.93 & 90.84 & 80.76 & 88.49 & 92.25 & 83.46 & 34.32 & 23.35 & 19.42 & 86.94 & 75.35 & 84.62 \\
 & SVG & 96.82 & 97.34 & 99.03 & 98.48 & 78.49 & 67.71 & \textbf{71.91} & \textbf{26.29} & 86.88 & 80.02 & 87.67 & 91.84 & 85.26 & \textbf{38.74} & 23.11 & 19.59 & 86.86 & \textbf{75.89} & 84.66 \\
 & SVG2 & 96.64 & 97.15 & \textbf{99.14} & \textbf{98.67} & 79.26 & 69.04 & 71.12 & 26.21 & 87.80 & 77.20 & 89.40 & 92.59 & 85.02 & 37.77 & \textbf{23.73} & 19.14 & 87.07 & 75.79 & \textbf{84.82} \\
 & SVOO & \textbf{96.91} & \textbf{98.30} & 98.83 & 98.03 & 79.24 & \textbf{70.57} & 71.09 & 23.61 & 86.21 & 75.80 & \textbf{90.86} & 91.48 & 86.86 & 36.38 & 23.41 & 19.14 & 87.12 & 74.62 & 84.62 \\
 & SVG-EAR & 96.38 & 97.52 & 98.77 & 98.38 & 81.57 & 70.25 & 70.98 & 25.56 & \textbf{92.86} & 81.72 & 87.60 & \textbf{93.81} & 84.34 & 31.78 & 23.14 & 19.11 & \textbf{87.14} & 75.51 & 84.81 \\
 & \proj & 96.65 & 97.84 & 98.86 & 98.44 & \textbf{83.62} & 67.86 & 71.19 & 23.85 & 88.74 & \textbf{82.53} & 89.60 & 91.31 & \textbf{87.67} & 33.44 & 23.26 & \textbf{19.75} & \textbf{87.14} & 75.45 & 84.80 \\
\bottomrule[0.15em]
\end{tabular}
}
\end{table*}

\proj attains the highest total score among sparse methods on HunyuanVideo, Wan2.1, Wan2.2, and LingBot, and is within 0.02 points of the best sparse method on LongCat-Video.
It also attains the highest quality score among sparse methods on Wan2.1, Wan2.2, and LingBot, and ties with SVG-EAR on LongCat-Video.
On every model, the total score of \proj stays within 0.05 points of the original model or exceeds it.
\section{Hyperparameters}
\label{sec:supp_hyperparameters}
\Tbl{tab:supp_hyper_main} lists the hyperparameters of the main evaluation for each model.
Full attention is kept for the first 10 denoising steps and the first 3\% of layers are kept dense for all methods.
The clustering configuration ($Q_C=256$, $K_C=1024$, two k-means iterations at initialization and per step) is shared by all clustering-based methods.
SVG is configured by its density, and SVG2, SVOO, and SVG-EAR by top-$p$ = 0.9.
\proj uses top-$p$ = 0.8.
Since \mode{SVG} does not incur clustering or the associated token-reordering overhead, equal attention density does not imply equal execution cost.
We therefore tune its sparsity parameter separately to account for these overhead differences when comparing quality--speed trade-offs.

\Tbl{tab:supp_hyper_sens} lists the settings of the density sweep in \Tbl{tab:sensitivity}.
Each row corresponds to one estimated density, and each cell gives the configuration of the method that reaches it.

\section{Additional Qualitative Results}
\label{sec:supp_qual}

\begin{figure*}[!t]
    \centering
    \includegraphics[width=0.93\textwidth]{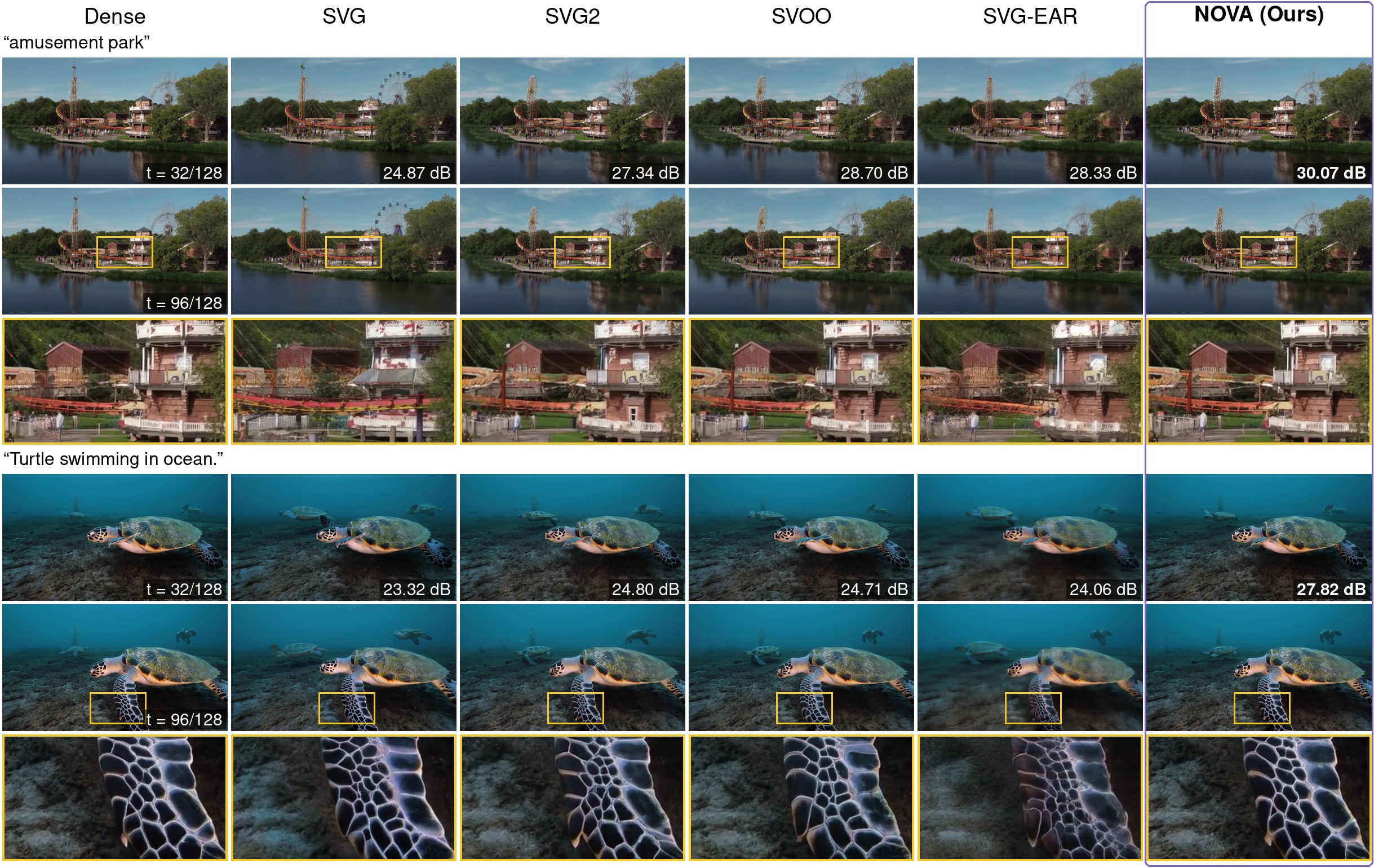}
    \caption{Qualitative comparison on \textbf{HunyuanVideo} ($720\times1280$, 129 frames).}
    \label{fig:supp_qual_hunyuan}
\end{figure*}

\begin{figure*}[p]
    \centering
    \includegraphics[width=0.93\textwidth]{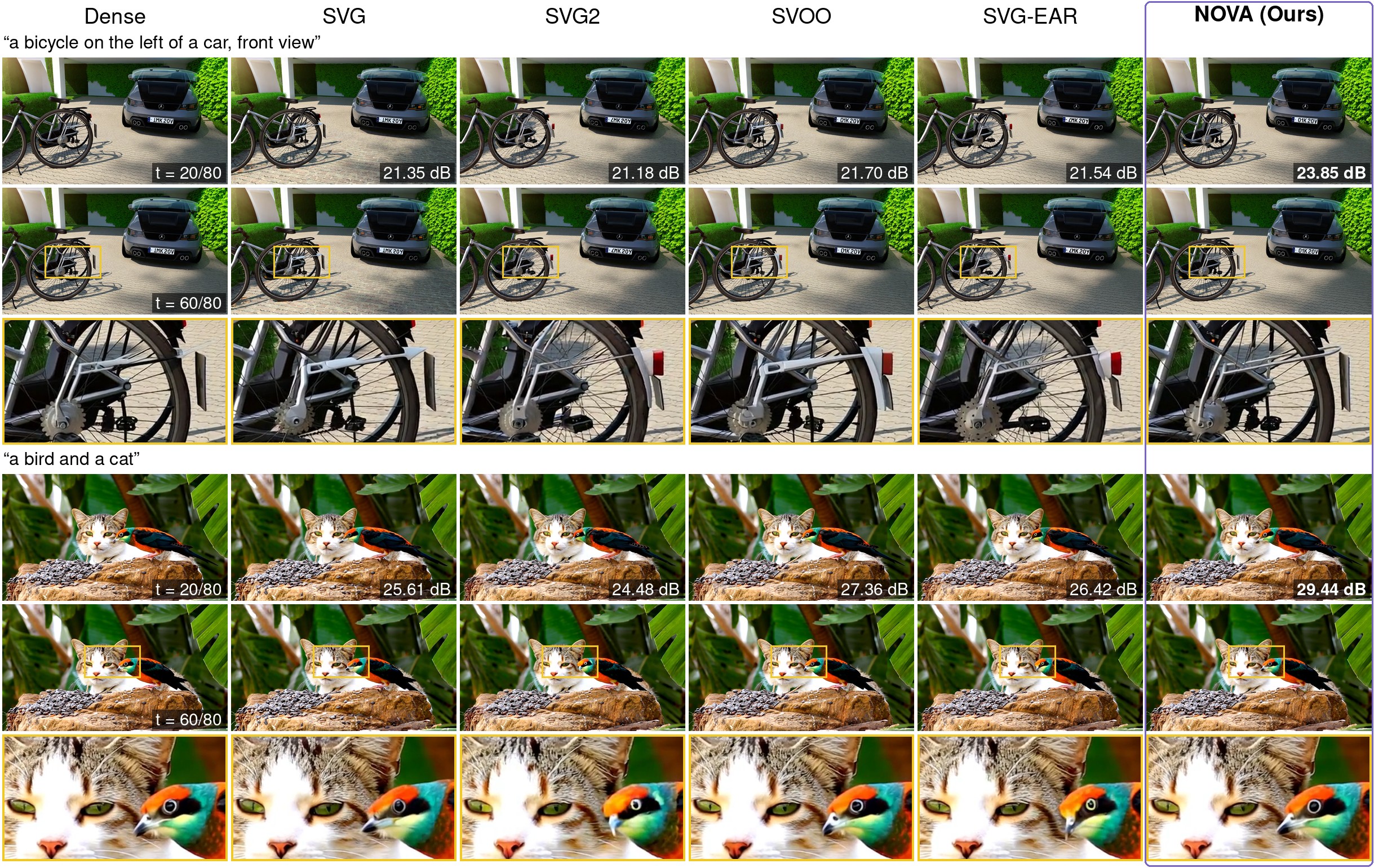}
    \caption{Qualitative comparison on \textbf{Wan2.1-T2V-14B} ($720\times1280$, 81 frames).}
    \label{fig:supp_qual_wan21}
\end{figure*}

\begin{figure*}[p]
    \centering
    \includegraphics[width=0.93\textwidth]{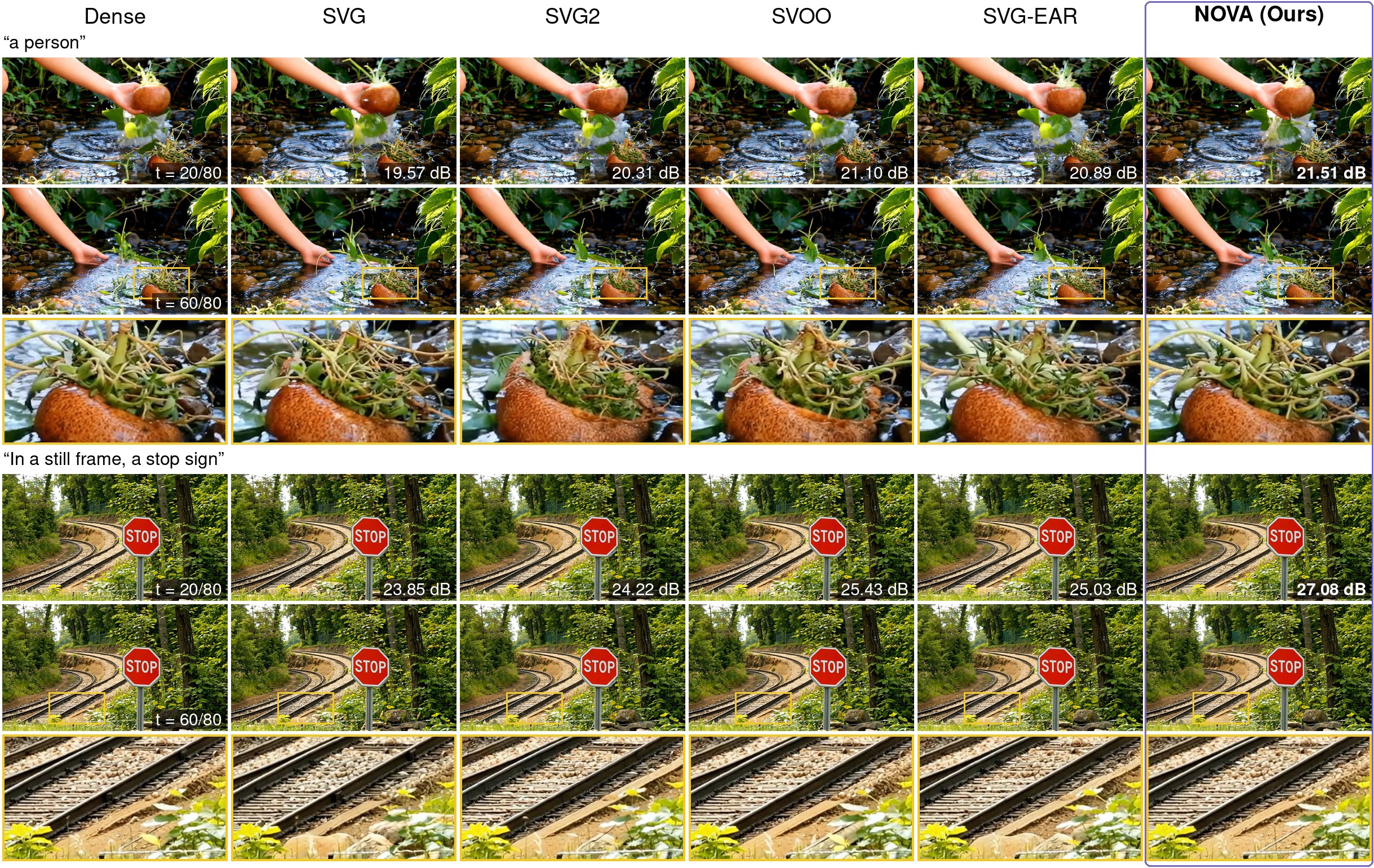}
    \caption{Qualitative comparison on \textbf{Wan2.2-T2V-A14B} ($720\times1280$, 81 frames).}
    \label{fig:supp_qual_wan22}
\end{figure*}

\begin{figure*}[p]
    \centering
    \includegraphics[width=0.93\textwidth]{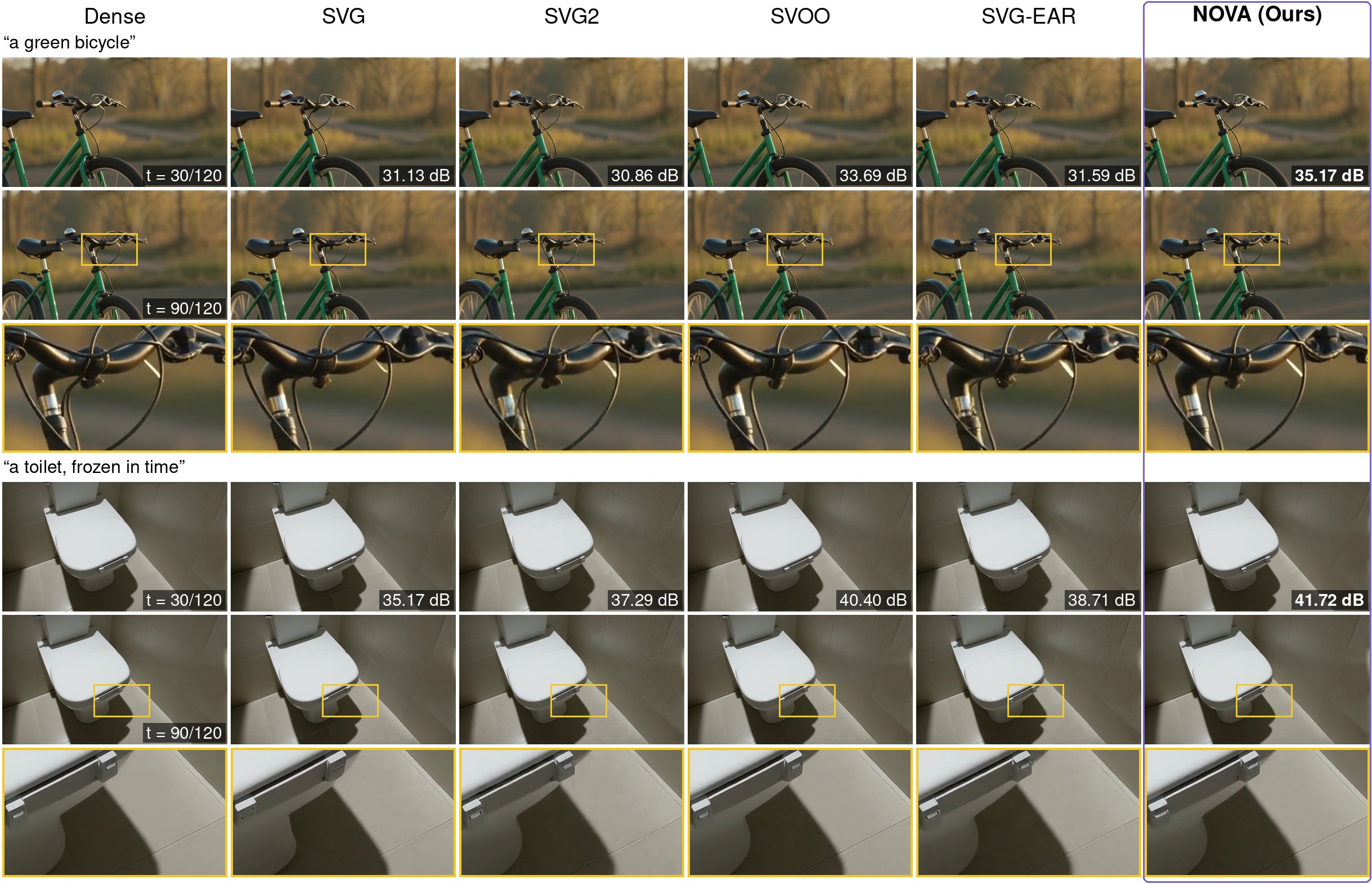}
    \caption{Qualitative comparison on \textbf{LingBot-Video-Dense-1.3B} ($736\times1280$, 121 frames).}
    \label{fig:supp_qual_lingbot}
\end{figure*}

\begin{figure*}[p]
    \centering
    \includegraphics[width=0.93\textwidth]{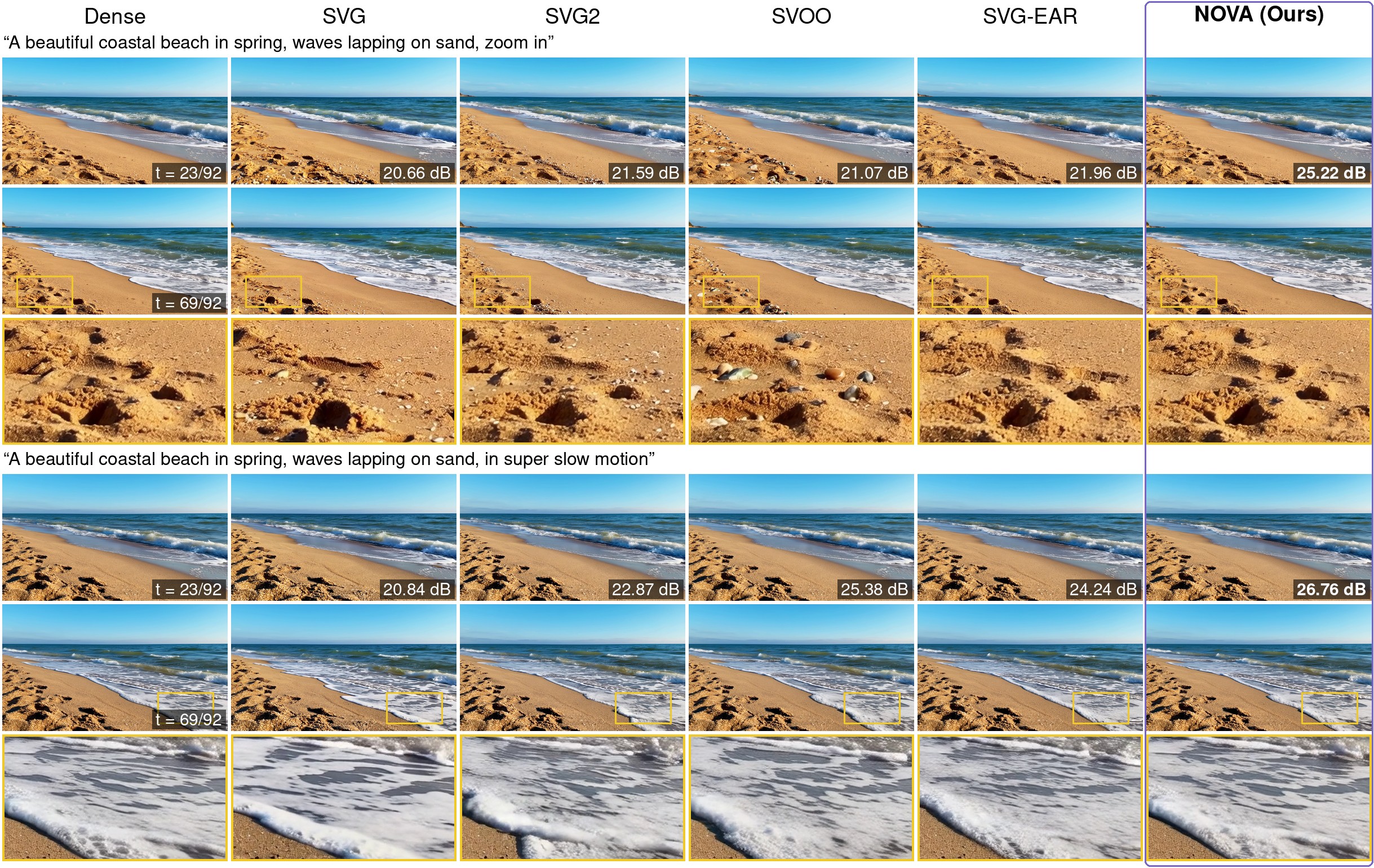}
    \caption{Qualitative comparison on \textbf{LongCat-Video} ($720\times1280$, 93 frames).}
    \label{fig:supp_qual_longcat}
\end{figure*}

\Fig{fig:supp_qual_hunyuan}--\Fig{fig:supp_qual_longcat} show video results for a subset of the evaluation prompts on all five models.
For each prompt, the first two rows show frames at one quarter and three quarters of the video, and the third row enlarges the yellow box in the second row.
The number on each frame is the PSNR of the whole video against Dense.
In the enlarged regions, \proj stays closest to the Dense video.

\end{document}